\documentclass[sigconf]{acmart}
\usepackage{enumitem}
\usepackage{bm}
\usepackage{multirow}
\usepackage{subfigure}
\usepackage{multirow}
\usepackage[normalem]{ulem}
\usepackage{balance}
\useunder{\uline}{\ul}{}

\usepackage{amsmath}
\usepackage{amsthm}
\usepackage{amsfonts}

\usepackage{cases}
\renewcommand{\shortauthors}{Chang Liu et al.}
\copyrightyear{2024}
\acmYear{2024}
\setcopyright{rightsretained}
\acmConference[KDD '24]{Proceedings of the 30th ACM SIGKDD Conference on Knowledge Discovery and Data Mining}{August 25--29, 2024}{Barcelona, Spain}
\acmBooktitle{Proceedings of the 30th ACM SIGKDD Conference on Knowledge Discovery and Data Mining (KDD '24), August 25--29, 2024, Barcelona, Spain}\acmDOI{10.1145/3637528.3671934}
\acmISBN{979-8-4007-0490-1/24/08}

\makeatletter
\gdef\@copyrightpermission{
  \begin{minipage}{0.3\columnwidth}
   \href{https://creativecommons.org/licenses/by/4.0/}{\includegraphics[width=0.90\textwidth]{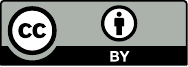}}
  \end{minipage}\hfill
  \begin{minipage}{0.7\columnwidth}
   \href{https://creativecommons.org/licenses/by/4.0/}{This work is licensed under a Creative Commons Attribution International 4.0 License.}
  \end{minipage}
  \vspace{5pt}
}
\makeatother

\begin{document}

\title{TDNetGen: Empowering Complex Network Resilience Prediction with Generative Augmentation of Topology and Dynamics}

\author{Chang Liu}

\affiliation{%
  \institution{Department of Electronic Engineering, \\
  BNRist, Tsinghua University}
  % \streetaddress{P.O. Box 1212}
  \city{Beijing}
  \country{China}
}
\email{lc23@mails.tsinghua.edu.cn}

\author{Jingtao Ding}
\authornote{Jingtao Ding and Yong Li are corresponding authors.}
\affiliation{%
  \institution{Department of Electronic Engineering, \\ BNRist, Tsinghua University}
  % \streetaddress{P.O. Box 1212}
  \city{Beijing}
  \country{China}
}
\email{dingjt15@tsinghua.org.cn}

\author{Yiwen Song}
\affiliation{%
  \institution{Shenzhen International Graduate School, \\ Tsinghua University}
  % \streetaddress{P.O. Box 1212}
  \city{Shenzhen}
  \state{Guangdong}
  \country{China}
}
\email{songyw23@mails.tsinghua.edu.cn}

\author{Yong Li}
\authornotemark[1]
\affiliation{%
  \institution{Department of Electronic Engineering, \\ BNRist, Tsinghua University}
  % \streetaddress{P.O. Box 1212}
  \city{Beijing}
  \country{China}
}
\email{liyong07@tsinghua.edu.cn}

% \affiliation{%

% \affiliation{%

% \affiliation{%

% \affiliation{%

% \affiliation{%

% \affiliation{%

% \affiliation{%

\renewcommand{\shortauthors}{Chang Liu, Jingtao Ding, Yiwen Song and Yong Li}
\newcommand{\rev}[1]{\textcolor{black}{#1}}
\begin{abstract}
Predicting the resilience of complex networks, which represents the ability to retain fundamental functionality amidst external perturbations or internal failures, plays a critical role in understanding and improving real-world complex systems. Traditional theoretical approaches grounded in nonlinear dynamical systems rely on prior knowledge of network dynamics. On the other hand, data-driven approaches frequently encounter the challenge of insufficient labeled data, a predicament commonly observed in real-world scenarios. In this paper, we introduce a novel resilience prediction framework for complex networks, designed to tackle this issue through generative data augmentation of network topology and dynamics. The core idea is the strategic utilization of the inherent joint distribution present in unlabeled network data, facilitating the learning process of the resilience predictor by illuminating the relationship between network topology and dynamics.
% The core idea is to benefit resilience predictor learning of the interplay between network topology and dynamics by exploiting the underlying joint distribution in unlabeled data.
Experiment results on three network datasets demonstrate that our proposed framework TDNetGen can achieve high prediction accuracy up to 85\%-95\%. Furthermore, the framework still demonstrates a pronounced augmentation capability in extreme low-data regimes, thereby underscoring its utility and robustness in enhancing the prediction of network resilience. We have open-sourced our code in the following link, \url{https://github.com/tsinghua-fib-lab/TDNetGen}.
\end{abstract}


\maketitle

\vspace{-0.3cm}
\section{Introduction}

\begin{figure}[t!]
    \centering
    \includegraphics[width=\linewidth]{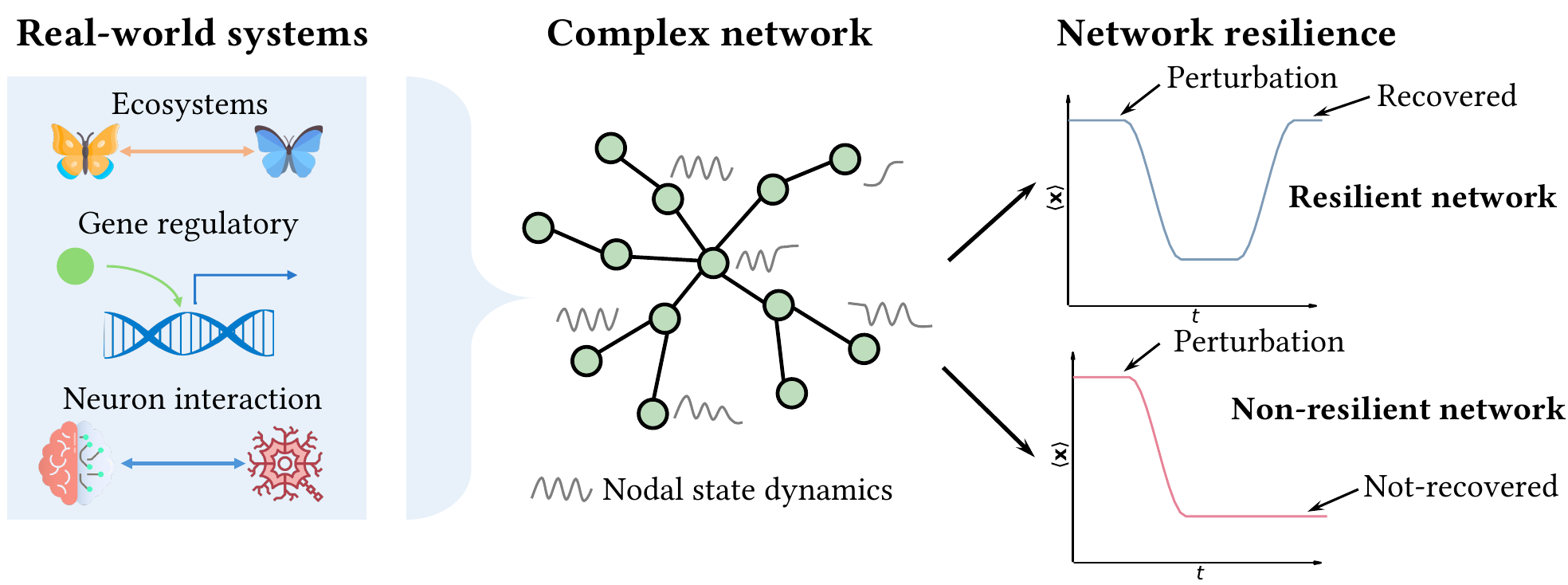}
    \vspace{-0.4cm}
    \caption{Resilience of complex networks. $\langle \mathbf{x}\rangle$ denotes the averaged nodal state of the network.}
    \label{fig:intro-resilience}
\vspace{-0.6cm}
\end{figure}

Real-world complex systems across various domains, such as ecological~\cite{holland2002population}, gene regulatory~\cite{alon2019introduction}, and neurological networks~\cite{wilson1972excitatory, wilson1973mathematical}, are often described as complex networks composed of interconnected nodes with weighted links. A fundamental characteristic of these systems is their resilience~\cite{may1977thresholds,gao2016universal}, that is, the ability to maintain functionality in the face of disruptions. From the perspective of dynamical systems, nodal state evolution of complex networks is driven by underlying nonlinear dynamics. Specifically, with the functionality of each node represented by its state value, a resilient network can recover from disruptions~(on its nodes) and dynamically evolve into a stable phase where all nodes operate at a high level of activity (see Figure~\ref{fig:intro-resilience}). Understanding and predicting this critical property of resilience in complex networks not only enhances our ability to analyze and intervene in natural and social systems~\cite{gao2016universal,zhang2022estimating,sanhedrai2022reviving,su2024rumor} but also offers valuable insights for the design of engineered infrastructures~\cite{xu2019resiliency}.

To predict network resilience, theories grounded in nonlinear dynamical systems have been developed~\cite{gao2016universal,laurence2019spectral,jiang2020inferring,zhang2020resilience}. 
These frameworks strive to separate the influences of network structure and dynamics to derive analytical solutions for complex, high-dimensional systems~\cite{ollerton2007finding,balaji2006comprehensive,gama2008regulondb}. However, \textit{theoretical approaches} often presuppose a detailed understanding of network evolution dynamics, which is usually not available in practical scenarios.
In contrast, \textit{data-driven methods} are capable of extracting both structural and dynamic information about networks directly from observational data~\cite{ding2024artificial,wang2023multi,mao2023detecting,mao2024identify,li2024skeleton}, allowing for resilience predictions without the need for predefined knowledge. From this perspective, the task of predicting network resilience can be reinterpreted as a \textit{graph classification problem} based on data of network structure and dynamics using machine learning techniques.
Nonetheless, the crucial role of resilience in system functionality means that collecting extensive labeled datasets from real-world complex networks is both expensive and impractical. As a result, the majority of network observations remain unlabeled, possessing information on network topology and nodal state trajectories but lacking resilience labels.

In this paper, we focus on addressing the problem of predicting network resilience amidst a scarcity of labeled data, identifying two primary obstacles: 

Firstly, designing models for resilience prediction is inherently complex due to the intricate interplay between network structure and dynamics. A network is considered resilient if it can consistently return to a state where all nodes are active following a prolonged period of self-evolution and neighborly interactions. However, while topological data is readily available, constructing a practical model requires the capability to make accurate predictions based on partial evolution trajectories collected from a short time window.

Secondly, enhancing prediction accuracy in the face of scarce labels involves leveraging the intrinsic information embedded in unlabeled data regarding network structure and dynamics. Existing methodologies mainly include pseudo-labeling~\cite{lee2013pseudo, tagasovska2019single,amini2020deep}, exemplified by self-training~\cite{iscen2019label}, and self-supervised learning~\cite{hu2019strategies, velivckovic2018deep, xu2021self,kim2022graph}. Pseudo-labeling tends to underperform with high model uncertainty, and self-supervised learning often overlooks the critical interplay between structure and dynamics, treating state evolution trajectories merely as node attributes. The graph data augmentation method~\cite{han2022g} emerges as a leading technique by utilizing unlabeled data distribution to generate diverse augmented samples for improved training. However, the challenge of comprehensively characterizing the distribution of both topology and dynamics in unlabeled data has yet to be tackled, especially with limited observations, such as a few labeled networks and incomplete evolution trajectories.

To fully resolve these challenges, we introduce a novel resilience prediction framework called TDNetGen, which utilizes generative augmentation of network topology and dynamics. The core of TDNetGen is a neural network-based predictor that integrates a graph convolutional network-based topology encoder together with a transformer-based trajectory encoder, capturing the complex relationship between network structure and dynamics. This predictor is further refined through training on an augmented dataset comprising resilient and non-resilient samples, \textit{i.e.}, networks with topology information and evolution trajectories. 

TDNetGen leverages a generative data augmentation approach by 1) capturing the underlying joint distribution of topology and dynamics in unlabeled data, and 2) obtaining the corresponding conditional distribution for each class label through a classifier-guided approach~\cite{dhariwal2021diffusion}. 
To facilitate effective generative learning in the vast joint space of topology and dynamics, we decouple the generation process into topology generation using a topology denoising diffusion module and dynamics simulation with a dynamics learning module. 
To ensure robust learning with limited observations, we incorporate a fine-tuning step for the resilience predictor on generated trajectories, thereby improving its generalization ability on unseen data.
To summarize, our main contributions are as follows.

\begin{itemize}[leftmargin=*]
    \item We tackle the critical problem of predicting complex network resilience under label sparsity issue and provide a novel perspective of improving by data augmentation.

    \item We design a generative augmentation framework that benefits resilience predictor learning of interplay between network topology and dynamics by exploiting the underlying joint distribution in unlabeled data.

    \item Empirical results on three network datasets demonstrate the superiority of our TDNetGen over state-of-the-art baselines in terms of increasing network resilience prediction accuracy up to 85\%-95\%. Moreover, aided by a generative learning capability of both topology and dynamics, TDNetGen can provide robust augmentation in low-data regimes, maintaining $98.3\%$ of performance even when dynamic information cannot be observed in unlabeled data.
\end{itemize}

\vspace{-0.35cm}
\begin{figure*}
    \centering
    \includegraphics[width=\linewidth]{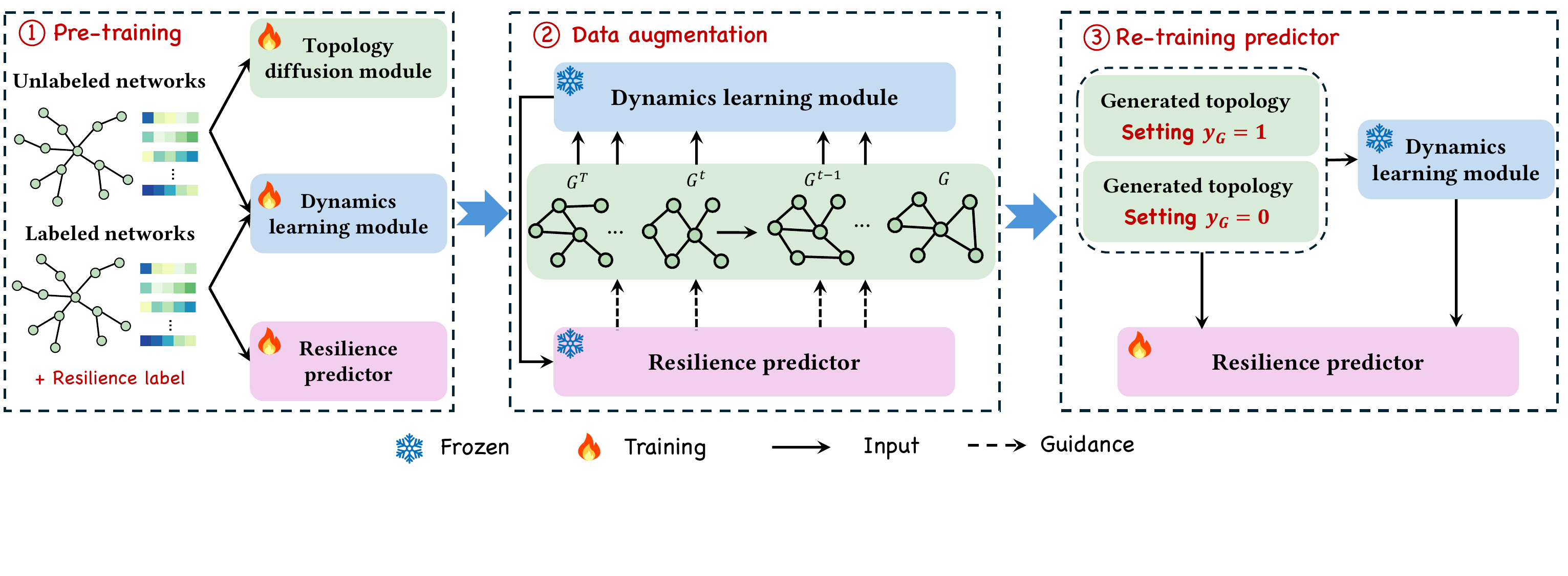}
    \vspace{-0.5cm}
    \caption{\rev{Overview of the proposed framework TDNetGen.}}
    \label{fig:overview}
    \vspace{-0.4cm}
\end{figure*}
\vspace{-0.1cm}
\section{Preliminaries}
\subsection{Resilience Prediction}\label{sec:resipred} 
Network resilience articulates that a resilient system is characterized by its invariable convergence towards a desired, non-trivial stable equilibrium following perturbation~\cite{gao2016universal}. Formally, given a complex network $G = (\mathbf{V}, 
\mathbf{A})$, where $\mathbf{V} = \{v_1, v_2, \cdots, v_N\}$ represents its node set and $\mathbf{A}$ denotes the adjacency matrix. The state of node $i$ can be represented as $x_i$, usually governed by the following non-linear ordinary differential equations (ODEs) as the nodal state dynamics:
\begin{equation}\label{equ:dynamics}
    \frac{dx_i}{dt}=F(x_i) + \sum_{j=1}^N A_{ij}G(x_i, x_j),
\end{equation}
where $F(x_i)$ represents the self-dynamics of nodes and $G(x_i, x_j)$ denotes interaction dynamics. The complex network $G$ is considered resilient if it consistently converges to only the desired nodal state equilibrium as time $t$ approaches infinity, irrespective of any perturbation and varying initial conditions with the exception of its fixed points.

\vspace{-0.3cm}
\subsection{Problem Formulation}
Considering the challenge of obtaining detailed knowledge of the underlying equations that govern nodal state dynamics in real-world scenarios, in this work, we advocate for a purely data-driven approach to predict network resilience.
In the context of the resilience prediction task, our dataset comprises network samples from which we can extract both topology and the initial $T$ steps of $M$ nodal state trajectories prior to reaching a steady state. 
Formally, for a network comprising $N$ nodes, the topology is represented by an adjacency matrix $\mathbf{A} \in \mathbb{R}^{N \times N}$, while the observed nodal state trajectories are denoted as $\mathbf{X} \in \mathbb{R}^{M \times N \times T}$. 
As demonstrated in Section~\ref{sec:resipred}, determining the resilience of a network precisely necessitates knowledge of its steady-state conditions, a requirement that is often prohibitive to meet due to the high observational costs (e.g., long-term species population growth). Consequently, only a limited subset of network samples are labeled, denoted as $\mathcal{P}$, with the majority remaining unlabeled, denoted as $\mathcal{Q}$, where $|\mathcal{P}|$ is significantly smaller than $|\mathcal{Q}|$. 
The reliance on a narrow labeled dataset $\mathcal{P}$ for training the resilience prediction model could result in sub-optimal performance due to the constrained sample size. In this work, we endeavor to leverage the untapped potential of the unlabeled data $\mathcal{Q}$ to enhance the training process of the resilience prediction model, with the objective of achieving superior predictive accuracy.

\vspace{-0.3cm}
\section{Methodology}
\subsection{Overview of Proposed Framework}

In this section, we propose an effective method named TDNetGen to address the problem of complex network resilience prediction with limited labeled data samples via generative augmentation of topology and dynamics.
Figure~\ref{fig:overview} illustrates the holistic design of TDNetGen, which consists of the following components:
\begin{itemize}[leftmargin=*]
\item \textbf{Topology diffusion module.} To facilitate resilience prediction performance and address the lack of labeled data, we design a diffusion module to model the distribution of unlabeled network topology. Therefore, we can sample new network topologies from the learned distribution.
\item \textbf{Dynamics learning module.} We propose a neural ODE~\cite{chen2018neural, zang2020neural}-based dynamics learning module to learn nodal state changes of networks from observed trajectories. It can simulate nodal state trajectories for the generated topologies from the topology diffusion module.
\item \textbf{Resilience predictor.} We design a resilience predictor empowered by Transformer and graph convolutional networks (GCNs), which jointly models nodal state dynamics and node interactions from observed trajectories and network topologies, respectively. It learns a low-dimensional embedding for each network and predicts its resilience based on this representation.
\end{itemize} 
\rev{In our proposed framework, we first train both the dynamics learning module and the topology diffusion module utilizing unlabeled as well as labeled nodal state trajectories and network topologies, respectively, which is then followed by the pre-training of the resilience predictor using accessible labeled data. Subsequently, we generate new samples facilitated by the topology diffusion module and dynamics learning module, with the guidance provided by the resilience predictor. The newly generated samples further enhance the training of the resilience predictor, thereby creating a synergistic feedback loop that significantly improves its predictive accuracy.}
\vspace{-0.3cm}
\subsection{Topology Diffusion Module}
\begin{figure}
    \centering
    \includegraphics[width=\linewidth]{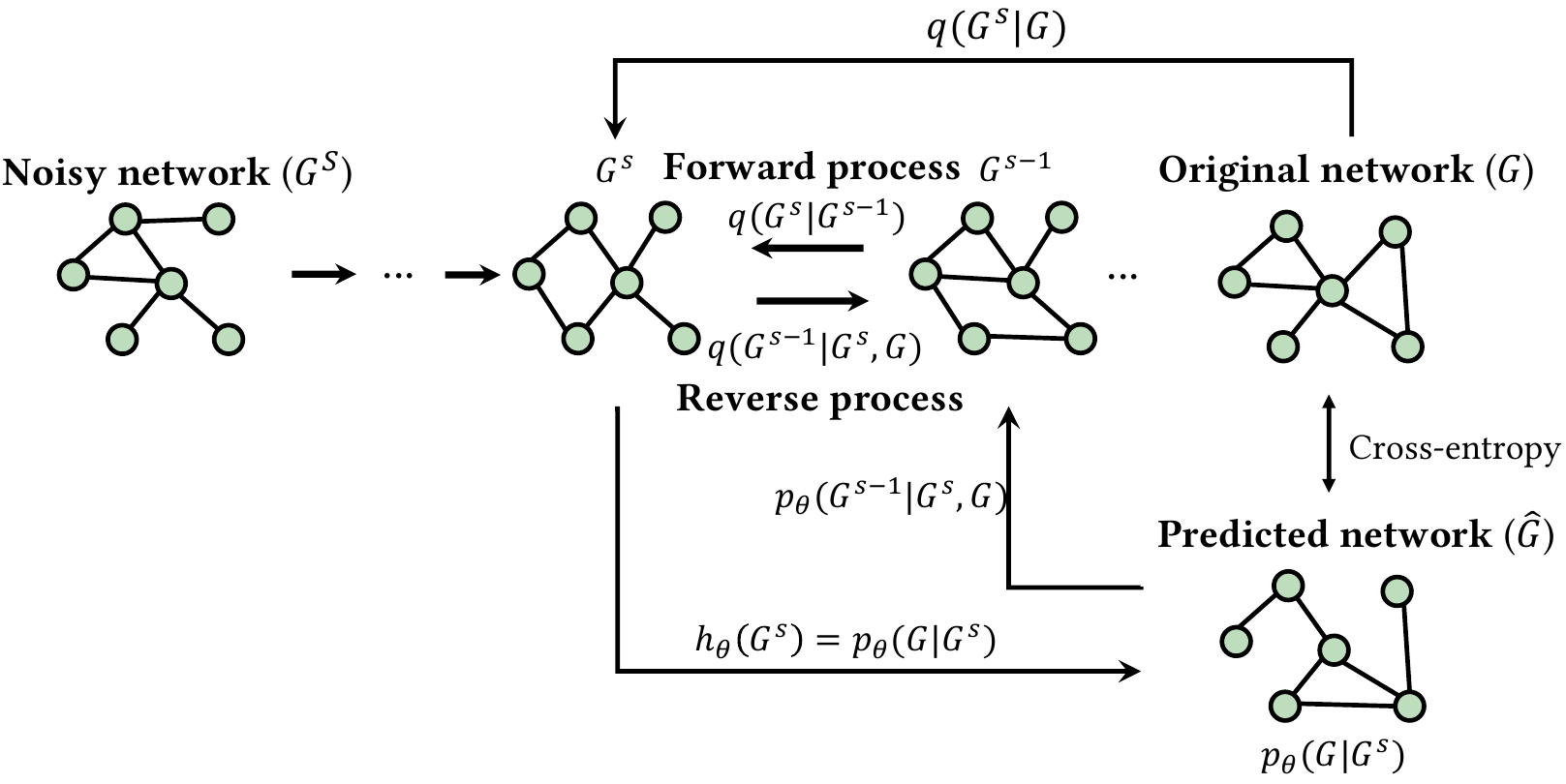}
    \vspace{-0.6cm}
    \caption{Illustration of topology diffusion module.}
    \label{fig:topo}
    \vspace{-0.5cm}
\end{figure}

\rev{Existing continuous graph diffusion models~\cite{niu2020permutation,jo2022score} undermine the sparsity nature of topologies and usually result in complete graphs lacking physically meaningful edges. Consequently, they fail to capture the structural properties of complex networks.} Therefore, we propose to model the distribution of network topologies using the discrete-space diffusion model~\cite{austin2021structured,vignac2022digress}, as illustrated in Figure~\ref{fig:topo}. Different from diffusion models for images with continuous Gaussian noise, here we apply a discrete type of noise on each edge, and the type of each edge can transition to another during the diffusion process. Here, we define the transition probabilities of all edges at time step $s$ as matrix $\mathbf{Q}^s$, where $\mathbf{Q}^s_{ij} = q(e^s=j|e^{s-1}=i)$ denotes the type of edge $e$ transits to $j$ from $i$ at time step $s$. The forward process of adding noise of each time step to graph structure $G$ is equivalent to sampling the edge type from the categorical distribution, formulated as:
\begin{gather}
    q(G^s|G^{s-1}) = \mathbf{E}^{s-1}\mathbf{Q}^s,
    q(G^s|G) = \mathbf{E}^{s-1}\bar{\mathbf{Q}}^s,
\end{gather}
where \rev{$\mathbf{E}\in \mathbb{R}^{N\times N\times 2}$ is the expanded adjacency matrix from $A$. Its last dimension is a 2-D one-hot vector where $[0, 1]$ denotes an edge exists between the corresponding nodes, while $[1,0]$ denotes there is no edge.} $\bar{\mathbf{Q}}^s = \mathbf{Q}^1\dots\mathbf{Q}^s$.
The reverse process aims to gradually recover the clean graph $G$ given a noisy graph $G^s$. Towards this end, inspired by existing works~\cite{austin2021structured,vignac2022digress}, we train a parameterized neural network $h_\theta$ which takes the noisy graph $G^s$ as input and predicts the structure of the clean graph $G$, \textit{i.e.,} all the probability $\hat{p}_{ij}$ of the existence of an edge $e_{ij}$ between node $i$ and $j$ in the clean graph $G$. We use the cross-entropy loss to optimize parameters $\theta$, formulated as follows:
\begin{equation}
    \mathcal{L}_{BCE}=\frac{1}{N^2}\sum_{1 \leq i, j \leq N}\text{CrossEntropy}(e_{ij}, \hat{p}_{ij}).
\end{equation}
For the parameterization of $h_\theta$, we employ the widely-recognized backbone of multi-layer graph transformers proposed by Dwivedi et al.~\cite{dwivedi2020generalization}. Intuitively, node features are updated in each layer through the self-attention mechanism, and edge features are updated from the information of its head and tail nodes. We describe the details of the parameterization network in Appendix~\ref{app:network}.

Once we train the neural network $h_\theta$, it can be applied to generate new network topologies. Specifically, the reverse process needs to estimate $p_\theta(G^{s-1}|G^s)$, which can be decomposed as follows, 
\begin{equation}\label{equ:decomp}
    p_\theta(G^{s-1}|G^s) = \prod_{1 \leq i, j \leq N} p_\theta(e_{ij}^{s-1}|G^s).
\end{equation}
Each term in Equ.~(\ref{equ:decomp}) can be formulated as,
\begin{equation}
p_\theta(e_{ij}^{s-1}|G^s) = \sum_{e \in \mathcal{E}}p_\theta(e_{ij}^{s-1}|e_{ij}=e, G^s)\hat{p}_{ij}(e),
\end{equation}
where 

\begin{equation}\label{equ:eij}
\small
    p_\theta\left(e_{ij}^{s-1} | e_{ij}=e, G^s\right)= \begin{cases}q\left(e_{ij}^{s-1} | e_{ij}=e, e_{ij}^s\right) & \text { if } q\left(e_{ij}^{s-1} | e_{ij}=e\right) >0 \\ 0 & \text { otherwise }\end{cases}
\end{equation}
can be calculated with Bayesian rule. After sampling for preset $S$ steps, we can generate new network topologies which follow the distribution of the training dataset.
\vspace{-0.4cm}
\subsection{Dynamics Learning Module}
Through the topology diffusion module, we can generate new network topologies for the training of resilience predictor. Nonetheless, we also need to obtain their nodal states trajectories to predict their resilience. As illustrated in Section~\ref{sec:resipred}, nodal state dynamics in complex networks usually have the generalized form of an ordinary differential equation (ODE) as:
\begin{equation}
    \rev{\frac{d\mathbf{x}(t)}{dt}=f(\mathbf{x}, G, \mathbf{W}, t),}
\end{equation}
where $\mathbf{x}(t) \in \mathbb{R}^{N}$ represents nodal states of $N$-nodes network $G$ at time step $t$, $f(\cdot)$ denotes the dynamics function, and \rev{$\mathbf{W}$ denotes all dynamics parameters}.
Therefore, we develop a dynamics learning module designed to infer changes in nodal states solely from data, which learns nodal state dynamics in the expressive hidden space based on neural-ODE~\cite{chen2018neural, zang2020neural}. 
Given the initial state $\mathbf{x}(0)$ of all network nodes, for each time step $t$, the process initiates by mapping the state of the nodes to a latent space through an encoder $f_e$. Subsequently, graph neural networks (GNNs) are utilized as a parameterization technique to facilitate the learning of dynamics within this latent space. The transition from latent space representation back to the nodal state at each time step is accomplished by employing a decoder function $f_d$, which decodes the hidden space embeddings to reconstruct the nodal states. The procedure can be represented as:
\begin{gather}
    \mathbf{x}_h(t) = f_e(\mathbf{x}(t)),\\
    \frac{d\mathbf{x}_h(t)}{dt} = \text{GNN}(\mathbf{x}_h(t)),\\
    \hat{\mathbf{x}}(t + \delta) =\mathbf{x}(t) + \int_{t}^{t+\delta}f_d(\frac{d\mathbf{x}_h(t)}{dt}),
\end{gather}
 where GNN can be implemented as an arbitrary design type of graph neural network layers. In our works, without the loss of generality, we choose to implement both encoder $f_e$ and decoder $f_d$ functions using MLPs. Furthermore, GNN is instantiated through graph convolutional networks~\cite{kipf2016semi}, thereby leveraging their robust capabilities in capturing and processing the inherent topological features of graphs.
We use $\ell_1$-loss to train the dynamics learning module, formulated as follows:
\begin{equation}\label{equ:l1loss}
    \mathcal{L}_1 = \frac{1}{|\mathcal{P}| + |\mathcal{Q}|}\sum_{i=1}^{|\mathcal{P}| + |\mathcal{Q}|}\int_0^T|\mathbf{x}_i(t)-\hat{\mathbf{x}_i}(t)|.
\end{equation}
As shown in Equ.~(\ref{equ:l1loss}), we train the dynamics learning module on both labeled dataset $\mathcal{P}$ and unlabeled dataset $\mathcal{Q}$ to achieve a better performance. It is noteworthy that in Section~\ref{sec:overall}, we demonstrate the dynamics learning module can also perform well even when the nodal states of unlabeled data are inaccessible.
\vspace{-0.3cm}
\subsection{Resilience Predictor}
\begin{figure}
    \centering
    \includegraphics[width=\linewidth]{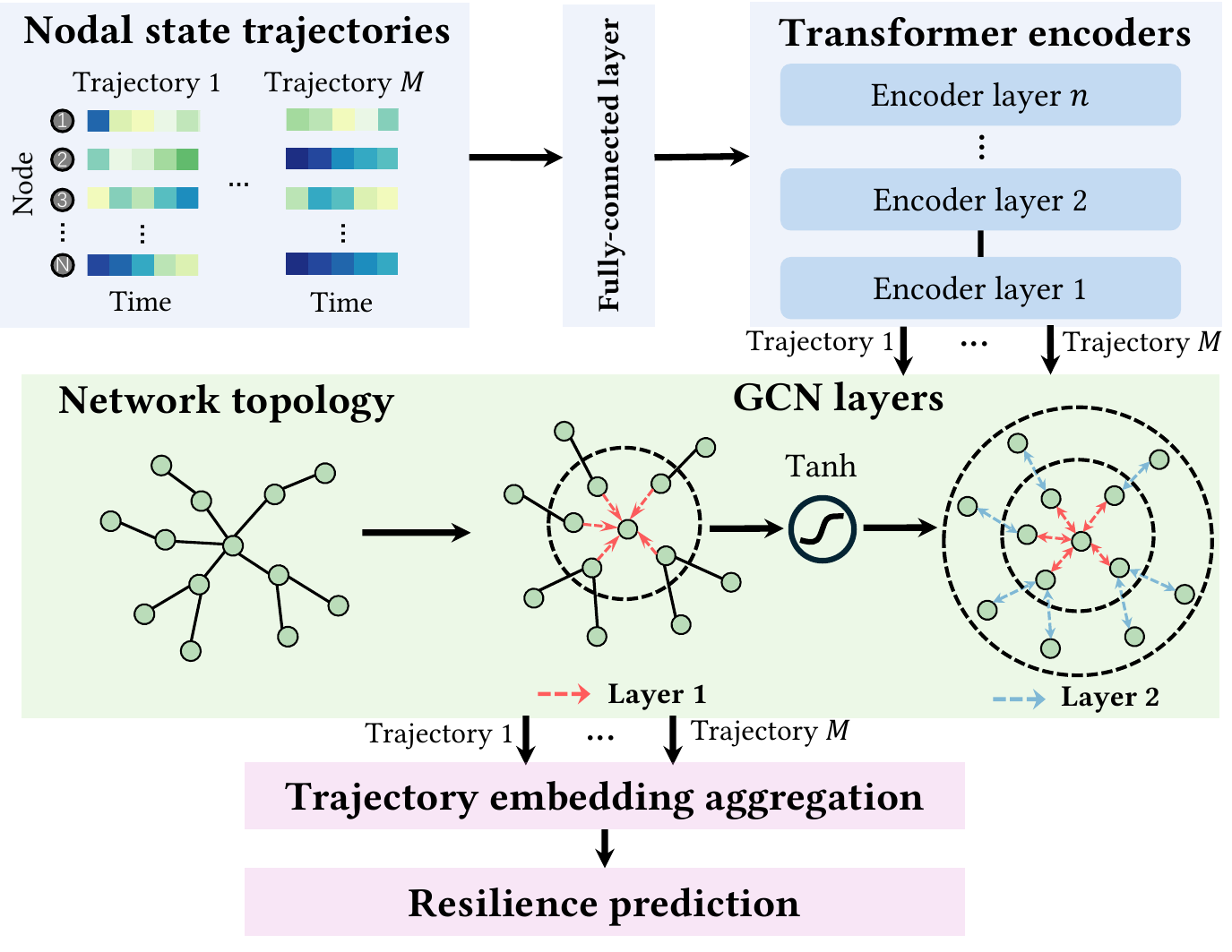}
    \vspace{-0.6cm}
    \caption{Architecture of the resilience predictor.}
    \label{fig:respred}
    \vspace{-0.6cm}
\end{figure}

We design a resilience predictor to jointly model the dynamics and topology of networks, which leverages stacked Transformer~\cite{vaswani2017attention} encoder layers and graph convolutional layers~\cite{kipf2016semi} to encode the temporal correlations of nodal states and learn spatial interactions within network topology, respectively. We illustrate its architecture in Figure~\ref{fig:respred}.
Specifically, for a network with $N$ nodes, we denote the nodal states with $d$ observed steps and $M$ trajectories of node $u$ as $\mathbf{x}_u \in \mathbb{R}^{d \times M}$. For the $k$-th trajectory $\mathbf{x}_{u, k} \in \mathbb{R}^{d}$, we first input its states of each time step to a feed-forward layer, and further encode the temporal correlation between time steps with Transformer encoder layers, formulated as follows, 
\begin{equation}
\mathbf{z}_{u,k} = \text{TransformerEncoder}(\mathbf{x}_{u,k}\mathbf{W}_1+\mathbf{b}_1),
\end{equation}
where $\mathbf{W}_1 \in \mathbb{R}^{1 \times d_e}$ and $\mathbf{b}_1 \in \mathbb{R}^{d_e}$ are trainable parameters.
After that, we integrate the embedding of the terminal time step of all nodes in the network, denoted by $\mathbf{Z}_k \in \mathbb{R}^{N \times d_e}$, as their $k$-th trajectory embeddings. 

To capture the interactions of nodes within the topology, we design a graph convolutional network (GCN) empowered by multi-layer message-passing. Given the adjacency matrix of network topology $\mathbf{A}$, we first calculate the Laplacian operator $\mathbf{\Psi} = \mathbf{I} - \mathbf{D}_{in}^{-\frac{1}{2}}\mathbf{A}\mathbf{D}_{out}^{-\frac{1}{2}}$, where the diagonal of $\mathbf{D}_{in}$ and $\mathbf{D}_{out}$ represent the in- and out-degree of nodes. We input the $k$-th trajectory embeddings of nodes to the graph convolutional network. The $l$-th layer message passing of the designed GCN can be represented as follows:
\begin{equation}
\mathbf{Z}_{k}^{(l)} = f(\mathbf{Z}_{k}^{(l-1)}) + g(\mathbf{\Psi}\mathbf{Z}_{k}^{(l-1)}),
\end{equation}
where $f(\cdot)$ and $g(\cdot)$ are implemented as MLPs. Such message-passing design is motivated by Equ.~(\ref{equ:dynamics}), aiming to more precisely model the effects from both the node itself and its neighborhood on a specific node. 

It is noteworthy that during the aforementioned procedure, different trajectories are processed in parallel. We further introduce a trajectory attention module to integrate the information from different trajectories for network-level representation. Specifically, we treat node embedding matrix of different trajectories after $L$ layers' message passing as a combination of feature maps, and denote the results after mean and max pooling as $\mathbf{Z}^{(L)}_{avg} \in \mathbb{R}^M$ and $\mathbf{Z}^{(L)}_{max} \in \mathbb{R}^M$, respectively. After that, we feed them into a shared MLP, add and activate the outputs to compute attention weights of trajectories, formulated as:
\begin{equation}
\mathbf{\alpha} = \sigma(\text{MLP}(\mathbf{Z}^{(L)}_{avg}) + \text{MLP}(\mathbf{Z}^{(L)}_{max})),
\end{equation}
where $\mathbf{\alpha} \in \mathbb{R}^M$, and $\sigma(\cdot)$ denotes the sigmoid activation function. Therefore, the fused node embedding matrix can be derived from:
\begin{equation}
\mathbf{Z} = \sum_{k=1}^M \mathbf{\alpha}_k \odot \rev{\mathbf{Z}^{(L)}_k},
\end{equation}
where $\mathbf{Z} \in \mathbb{R}^{N \times d_e}$, $\alpha_k$ is the attention weight for the $k$-th trajectory, and $\odot$ denotes Hadamard product.
We use a readout function to derive the embedding of the entire network, \textit{i.e.,} $\mathbf{e}_{net} = \text{Readout}(\mathbf{Z})$. Here, we implement the readout function as mean pooling between nodes. We then predict the resilience $\hat{y}$ of the network using $\mathbf{e}_{net}$ as follows:
\begin{equation}
\hat{y} = \text{MLP}(\mathbf{e}_{net}).
\end{equation}
Then we can train the resilience predictor with binary cross-entropy (BCE) loss, 
\begin{equation}
    \mathcal{L}_{\text{BCE}} = \sum_{i=1}^{|\mathcal{N}|}y_i\log(\hat{y}_i) + (1-y_i)\log(1-\hat{y}_i),
\end{equation}
where $y_i$ and $y$ denote the ground truth and the prediction result of the $i$-th network. $|\mathcal{N}|$ is the number of networks used for training. However, its predictive performance typically falls below the optimal level, primarily attributed to the scarcity of data. 

It is noteworthy that after training resilience predictor on labeled data, we further fine-tune the predictor on identical topologies wherein the nodal state trajectories are generated through the neural-ODE of the dynamics learning module. It enables the resilience predictor to accurately accommodate the minor discrepancies observed between the ground-truth trajectories and those generated through simulation, thereby ensuring the robust predictive performance.

\vspace{-0.3cm}
\subsection{Joint Data Augmentation of Topology and Dynamics }
The above modules enable us to generate network samples with both topology and nodal state trajectories. However, it is important to note that the simulated nodal states are confined to the initial temporal period, corresponding to the maximal duration present within the training dataset, and compelling the dynamics learning module to simulate time steps beyond its training scope yields results of questionable reliability. Consequently, the principal challenge arises from the inability to ascertain the steady-state conditions of the generated networks. This limitation obstructs the direct acquisition of resilience labels, presenting a significant impediment to the data augmentation. 
To overcome this problem, we advocate for the strategy of guiding the topology diffusion module, enabling it to generate networks with predefined resilience characteristics. More precisely, we integrate classifier guidance~\cite{dhariwal2021diffusion} into the topology diffusion model, which leverages signals derived from the resilience predictor trained on the labeled dataset. The conceptual basis of the guidance mechanism involves that the resilience predictor provides the resilience condition of the clean samples from the intermediate samples generated by the diffusion model, which in turn, steers the generation process towards exhibiting desired resilience characteristics. To formally define the guided diffusion process, we provide the following lemma from~\cite{dhariwal2021diffusion}:
\begin{lemma}
    Denote the forward process conditioned on $y_G$ as $\hat{q}$, and the unconditional forward process as $q$. Given the reasonable assumption $\hat{q}(G^s|G, y_G) = q(G^s|G)$, we have 
\begin{equation}
\hat{q}(G^{s-1}|G^s, y_G) \propto q(G^{s-1}|G^s)\hat{q}(y_G|G^{s-1})
\end{equation}
\end{lemma}
An direct estimation of $q(G^{s-1}|G^s)\hat{q}(y_G|G^{s-1})$ is to use
\begin{equation}
p_\theta(G^{s-1}|G^s)p_\eta(y_G|G^{s-1}),
\end{equation}
where $p_\eta$ are parameterized by the resilience predictor. However, we cannot evaluate all possible values of $G^{s-1}$. A viable method is to treat $G$ as a continuous tensor of order $N^2$, and use the first-order approximation from Taylor expansion~\cite{vignac2022digress}, as
\begin{align}
\log\hat{q}(y_G|G^{s-1}) &\approx \log \hat{q}(y_G|G^s) + \left\langle\nabla_G \log \hat{q}\left(y_G \mid G^s\right), G^{s-1}-G^s\right\rangle \\
&\approx \sum_{1 \leq i, j \leq N} \left\langle \nabla_{e_{ij}}\log \hat{q}\left(y_G \mid G^s\right), e_{ij}^{s-1}\right\rangle + C(G^s),
\end{align}
where $C(G^s)$ is a function that only relates to $G^s$. Assume that $\hat{q}(y_G|G^s) \sim \text{Bernoulli}(f_\eta(G_s))$, where $f_\eta$ is the resilience predictor, we have
\begin{equation}
\nabla_{G^s} \log \hat{q}_\eta\left(y \mid G^s\right) \propto-\nabla_{G^s}\mathcal{L}_{BCE}(\hat{y}, y_G)
\end{equation}

Drawing upon the aforementioned theoretical framework, at the step $s$ of the reverse process, we first employ the resilience predictor $f_\eta$ to predict $y_G$, \textit{i.e.,} $\hat{y}_G = f_\eta(G^s)$, and estimate the $p_\eta(y_G|G^{s-1})$ as
\begin{equation} \label{equ:guide-intensity}
p_\eta(y_G|G^{s-1}) \propto \exp(-\lambda \langle\nabla_{G^s}\mathcal{L}_{BCE}(\hat{y}_G, y_G), G^{s-1}\rangle),
\end{equation}
where $\lambda$ represents the guidance intensity. Hence, we can sample $G^{s-1}$ from 
\begin{equation}
G^{s-1} \sim p_\theta(G^{s-1}|G^s)p_\eta(y_G|G^{s-1}),
\end{equation}
where $p_\theta(G^{s-1}|G^s)$ can be calculate from Equ.~(\ref{equ:decomp})-(\ref{equ:eij}).

Consequently, by setting $y_G = 1$ and $y_G = 0$, we can generate novel labeled network topologies guided by the resilience predictor. These topologies subsequently serve as inputs to simulate their respective nodal state trajectories via the dynamics learning module. This approach facilitates the augmentation of our datasets with additional fully labeled data, which, in turn, allows for the re-training of the resilience predictor. Such a method is anticipated to significantly enhance the predictive accuracy of the resilience predictor, ensuring a more reliable assessment of network resilience under conditions of data sparsity.
\vspace{-0.3cm}
\subsection{\rev{Time Complexity Analysis}}
\rev{We define $N$ as the number of nodes in a graph, and analyze time complexity of each module in our framework as follows.
\begin{itemize}[leftmargin=*,partopsep=0pt,topsep=0pt]
\item \textbf{Topology diffusion module} is parameterized using GraphTransformer layers (Appendix~\ref{app:network}). It exhibits a time complexity of $\mathcal{O}(N^2)$ per layer, attributable to the computation of attention scores and the prediction process for each edge.
\item \textbf{Dynamics learning module} is based on neural-ODE and parameterized through GCN layers. This module also demonstrates a time complexity of $\mathcal{O}(N^2)$ resulting from convolution operations and the application of a fourth-order Runge-Kutta ODE solver.
\item \textbf{Resilience predictor} leverages stacked Transformer encoder layers to capture temporal correlations among nodal states, while spatial interactions within the network topology are discerned through GCN layers. Time complexities of the Transformer encoder layers and GCN layers are $\mathcal{O}(NT^2)$ and $\mathcal{O}(N^2)$, respectively, with $T$ representing the trajectory length. Typically, $T$ is significantly smaller than $N$ for most graph structures.
\end{itemize}
Consequently, the overall time complexity of TDNetGen is dominantly $\mathcal{O}(N^2)$, signifying its scalability and efficiency in processing large graph structures. In practical experiments, our framework takes about 10 seconds to generate a 200-nodes graph with nodal state trajectories and 20 milliseconds to predict its resilience. Since resilience inference is not a real-time task, such time complexity is acceptable for application. }
\vspace{-0.3cm}
\section{Experiments}
In this section, we demonstrate the superior performance of our framework TDNetGen, aiming to answer the following research questions:
\begin{itemize}[leftmargin=*,partopsep=0pt,topsep=0pt]
\item \textbf{RQ1:} How does our framework TDNetGen compare to potential baseline methods of harnessing unlabeled data to enhance predictive performance?
\item \textbf{RQ2:} How do different designs of TDNetGen affect the model performance?
\item \textbf{RQ3:} How does TDNetGen perform across limited numbers of original labeled samples and lengths of nodal state trajectories?
\item \textbf{RQ4:} How does TDNetGen perform with different network types and scales?
\end{itemize}
\vspace{-0.3cm}
\subsection{Experimental Settings}\label{subsec:expsetting}
\subsubsection{Dataset}
\begin{table}[t]
\caption{Statistics of network datasets.}
\vspace{-0.4cm}
\label{tab:statistics}
\resizebox{\linewidth}{!}{
\begin{tabular}{cccc}
\hline
                     & \textbf{Mutualistic} & \textbf{Regulatory} & \textbf{Neuronal} \\ \hline \hline
\#Unlabeled networks & 1900        & 1900       & 1900     \\
\#Labeled networks   & 100          & 100         & 100       \\
Average \#nodes      & 36          & 44         & 45       \\
Average \#edges      & 99          & 115        & 112      \\ \hline
\end{tabular}
}
\vspace{-0.6cm}
\end{table}
% \usepackage{multirow}
% \usepackage[normalem]{ulem}
\begin{table*}[t]
\caption{Overall predictive performance of models. w/o trajectories represents that nodal state trajectories of unlabeled data are unknown. The performance of TDNetGen is marked in bold, and the best baseline is \ul{underlined}.}
\vspace{-0.4cm}
\label{tab:overall}
\tabcolsep=0.47cm
\begin{tabular}{cc|cc|cc|cc}
\hline
 &  & \multicolumn{2}{c|}{\textbf{Mutualistic}} & \multicolumn{2}{c|}{\textbf{Regulatory}} & \multicolumn{2}{c}{\textbf{Neuronal}} \\
\multicolumn{2}{c|}{\# Training samples} & \multicolumn{2}{c|}{100} & \multicolumn{2}{c|}{100} & \multicolumn{2}{c}{100} \\ \hline
\multicolumn{2}{c|}{\textbf{Model}} & \textbf{F1} & \textbf{ACC} & \textbf{F1} & \textbf{ACC} & \textbf{F1} & \textbf{ACC} \\ \hline
\multicolumn{2}{c|}{Vanilla model} & 0.838 & 0.848 & 0.806 & 0.780 & 0.775 & 0.784 \\ \hline
Self-training & ST & 0.807 & 0.827 & 0.780 & 0.735 & 0.728 & 0.764 \\ \hline
\multirow{6}{*}{\begin{tabular}[c]{@{}c@{}}Self-supervised  learning\end{tabular}} & EdgePred~\cite{hu2019strategies} & 0.840 & 0.851 & 0.813 & 0.791 & 0.776 & 0.784 \\
 & AttrMask~\cite{hu2019strategies} & 0.831 & 0.845 & 0.817 & 0.793 & 0.770 & 0.779 \\
 & ContextPred~\cite{hu2019strategies} & 0.843 & 0.847 & 0.815 & 0.789 & 0.772 & 0.781 \\
 & InfoMax~\cite{velivckovic2018deep} & 0.829 & 0.815 & 0.875 & 0.870 & 0.787 & 0.805 \\
 & GraphLog~\cite{xu2021self} & 0.808 & 0.796 & 0.796 & 0.769 & 0.772 & 0.732 \\
 & D-SLA~\cite{kim2022graph} & 0.810 & 0.799 & 0.855 & 0.840 & 0.780 & 0.805 \\ \hline
\multirow{2}{*}{Graph data augmentation} & TRY~\cite{gao2016universal} & {\ul 0.891} & 0.886 & 0.896 & 0.898 & {\ul 0.818} & {\ul 0.833} \\
 & G-Mixup~\cite{han2022g} & 0.875 & {\ul 0.888} & {\ul 0.900} & {\ul 0.899} & 0.786 & 0.812 \\ \hline
\multicolumn{2}{c|}{\textbf{TDNetGen (w/o trajectories)}} & \textbf{0.913} & \textbf{0.913} & \textbf{0.922} & \textbf{0.923} & \textbf{0.805} & \textbf{0.810} \\ 
\multicolumn{2}{c|}{\textbf{TDNetGen}} & \textbf{0.929} & \textbf{0.934} & \textbf{0.944} & \textbf{0.946} & \textbf{0.845} & \textbf{0.873} \\ \hline
\multicolumn{2}{c|}{\textbf{Improvement}} & \textbf{4.26\%} & \textbf{5.18\%} & \textbf{4.89\%} & \textbf{5.23\%} & \textbf{3.30\%} & \textbf{4.80\%} \\ \hline
\end{tabular}
\vspace{-0.4cm}
\end{table*}
To construct the dataset, we synthesize complex networks with three nodal state dynamics from physics and life sciences. Denote $x_i(t)$ as the state of node $i$ at time step $t$, the dynamics are as follows,
\begin{itemize}[leftmargin=*,partopsep=0pt,topsep=0pt]
\item \textbf{Mutualistic dynamics.} The mutualistic dynamics~\cite{holland2002population} $\frac{d x_i}{d t}=B+x_i\left(1-\frac{x_i}{K}\right)\left(\frac{x_i}{C}-1\right)+\sum_{j=1}^N A_{ij} \frac{x_i x_j}{D+E x_i+H x_j}$ describes the alterations in species populations that are engendered by the migration term $B$, logistic growth term with environment capacity $K$~\cite{zang2018power}, Allee effect~\cite{allee1949principles} term with threshold $C$, and mutualistic interaction between species with interaction network $\mathbf{A}$.
\item \textbf{Regulatory dynamics.} The regulatory dynamics, also called Michaelis-Menten dynamics~\cite{alon2019introduction}, is described by $\frac{d x_i}{d t}=-B x_i^f+\sum_{j=1}^N A_{i j} \frac{x_j^h}{x_j^h+1}$. $f$ represents the degradation ($f=1$)  or dimerization ($f=2$). Additionally, the second term in the equation is designed to capture genetic activation with Hill coefficient $h$, which serves to quantify the extent of gene regulation collaboration.
\item \textbf{Neuronal dynamics.} The neuronal dynamics, also called Wilson-Cowan dynamics~\cite{wilson1972excitatory,wilson1973mathematical}, is described by the equation of $\frac{d x_i}{d t}=-x_i+\sum_{j=1}^N A_{i j} \frac{1}{1+e^{\mu-\delta x_j}}$. For each node in the network, it receives cumulative inputs from its neighbors. The second term of the equation represents the activation signal that is collectively contributed by all neighboring nodes.
\end{itemize}
For each dynamics, we synthesize Erd\H{o}s-R\'enyi networks~\cite{erdHos1960evolution} with edge creation probability uniformly sampled in $[0,0.15]$, and use the fourth-order Runge-Kutta stepper~\cite{dormand1980family} to simulate their nodal state trajectories. For more details, please refer to the Appendix~\ref{app:data}. 
We create $2000$ network samples for training, $200$ for validation, and another $200$ samples for testing. In the training stage, we randomly select $100$ (5\%) samples as labeled data and keep other $1900$ samples as unlabeled. The statistics of datasets are shown in Table~\ref{tab:statistics}.
\subsubsection{Baselines and metrics}
In the following parts, we define the model trained only on original labeled data as the \textit{vanilla model}. Besides this, there are mainly three kinds of baseline methods designed to leverage unlabeled data for enhancing the predictive performance of the vanilla predictor.
\begin{itemize}[leftmargin=*,partopsep=0pt,topsep=0pt]
\item \textbf{Self-training methods.} They utilize the predictor to assign pseudo labels to unlabeled data, thereby augmenting the labeled training dataset. We abbreviate these method as ST.
\item \textbf{Self-supervised learning methods.} They employ hand-crafted tasks to derive insights from unlabeled data, thereby facilitating the pre-training of model parameters. Subsequently, they undergo further supervised training on the labeled dataset.
This approach is predicated on the premise that integrating pre-training phases with subsequent supervised learning phases leverages both unlabeled and labeled datasets, thereby enhancing the model's learning efficacy and predictive accuracy. Such methods include EdgePred, AttrMask, ContextPred~\cite{hu2019strategies}, InfoMax~\cite{velivckovic2018deep}, GraphLog~\cite{xu2021self}, and D-SLA~\cite{kim2022graph}.
\item \textbf{Graph data augmentation (GDA) methods.} They incorporate new graphs with labels to train the model, including theory-guided method (TRY~\cite{gao2016universal}, detailed in Appendix~\ref{app:baseline}) and G-Mixup~\cite{han2022g}.
\end{itemize}

\noindent For both self-training and graph data augmentation methods, the quantity of newly generated samples is the same as that produced by our method. Similarly, within the realm of self-supervised learning, we also select the same count of unlabeled samples as the volume of new samples generated by our framework. 
We use F1-score (F1) and Accuracy (ACC) to evaluate the predictive performance of the resilience predictor.
\subsubsection{Implementation details}
We implement our model in PyTorch and complete all training and test tasks on a single NVIDIA RTX 4090 GPU. With our framework, we generate $1000$ new networks assigning 500 resilient and 500 non-resilient networks. Subsequently, we randomly select half of these networks to serve as the augmented data. We set the guidance intensity $\lambda = 2000$.
In our study, model parameters are optimized using the Adam optimizer, coupled with Xavier initialization, which ensures a robust starting point for learning.
For each experiment, we conduct a minimum of at least 5 times employing distinct random seeds and report the average value.
\subsection{Overall Performance (RQ1)}\label{sec:overall}
We report the performance of our framework with mean value and standard deviation in Table~\ref{tab:overall}.
From the experimental results, we have the following conclusions:
\begin{itemize}[leftmargin=*,partopsep=0pt,topsep=0pt]
\item \textbf{Our framework effectively empowers predictive performance via generative augmentation of both topology and dynamics.}
The results demonstrate that with the help of our proposed data augmentation framework, the predictive performance of the resilience predictor can be effectively improved. For example, on mutualistic dataset, the F1-score of the resilience predictor previously trained on 100 labeled data increases from 0.838 to 0.929 (+10.86\%), and its ACC increases from 0.848 to 0.934 (+10.14\%) after training on the augmented data. Moreover, our framework also improves the best baseline among all self-training, self-supervised learning, and GDA methods \textit{w.r.t.} F1-score by $4.26\%$, $4.89\%$, $3.30\%$, and \textit{w.r.t.} ACC by $5.18\%$, $5.23\%$, $4.80\%$, on mutualistic, regulatory and neuronal dataset, respectively. All these results demonstrate the outstanding performance of our proposed framework. We find that the best baseline methods on three datasets belong to the category of graph data augmentation. Compared with TRY and G-mixup, we achieve to jointly model network topology and dynamics in a fine-grained manner. 
\item \textbf{Robustness performance without nodal state trajectories of unlabeled data.}
In certain contexts, the requirement to obtain the nodal states, even only for an initial phase of evolution, still proves to be difficult or costly. Consequently, we analyze scenarios where the nodal state trajectories of unlabeled data are inaccessible and we can only train the dynamics learning module on those of limited labeled data $\mathcal{P}$ in Equ.~(\ref{equ:l1loss}). Results in Table~\ref{tab:overall} demonstrate that our framework is capable of sustaining commendable performance even under such constrained conditions and surpassing the best baseline in most scenarios. It underscores the versatility of our framework and its potential effectiveness under more limited data availability of the real world scenarios.
\item \textbf{Self-training cannot universally guarantee a positive impact on model performance.} The results demonstrate that self-training methods have a relatively small positive effect on predictive performance among all three datasets compared to our framework TDNetGen. For example, on regulatory datasets, the F1-score of the resilience predictor increases 0.051, and its ACC increases 0.079 compared to the vanilla model. This is because the labels assigned to the augmented data in the self-training process originate from the model itself with sub-optimal predictive performance. This approach inherently carries the risk of generating labels that are incongruent with the ground truth and partially introduce contradictory information into the training dataset. The presence of such inaccurately labeled data can confound the learning algorithm, leading to a deterioration in the model's capacity to make accurate predictions.
\item \textbf{Extracting knowledge from unlabeled data via hand-crafted self-supervised tasks offers marginal benefits to the resilience prediction.}
We also find that models trained on self-supervised tasks can only extract limited knowledge from unlabeled data to benefit the resilience prediction task. From the results, the improvement to vanilla model from competitive self-supervised learning methods (ContextPred~\cite{hu2019strategies} and Infomax~\cite{velivckovic2018deep}) is still relatively marginal compared to our framework (+$10.86\%$, +$17.12\%$ and +$9.03\%$, on mutualistic, regulatory, neuronal dataset, respectively).
The primary reason for the observed discrepancy lies in the substantial divergence between conventional hand-crafted tasks, which only focus on the modeling of topological structures. In the resilience prediction task, however, we also need to consider nodal state dynamics of networks.

\end{itemize}
\subsection{Ablation Study (RQ2)}
\begin{figure}[t!]
    \centering
    \subfigure[F1-score]{
    \includegraphics[width=0.475\linewidth]{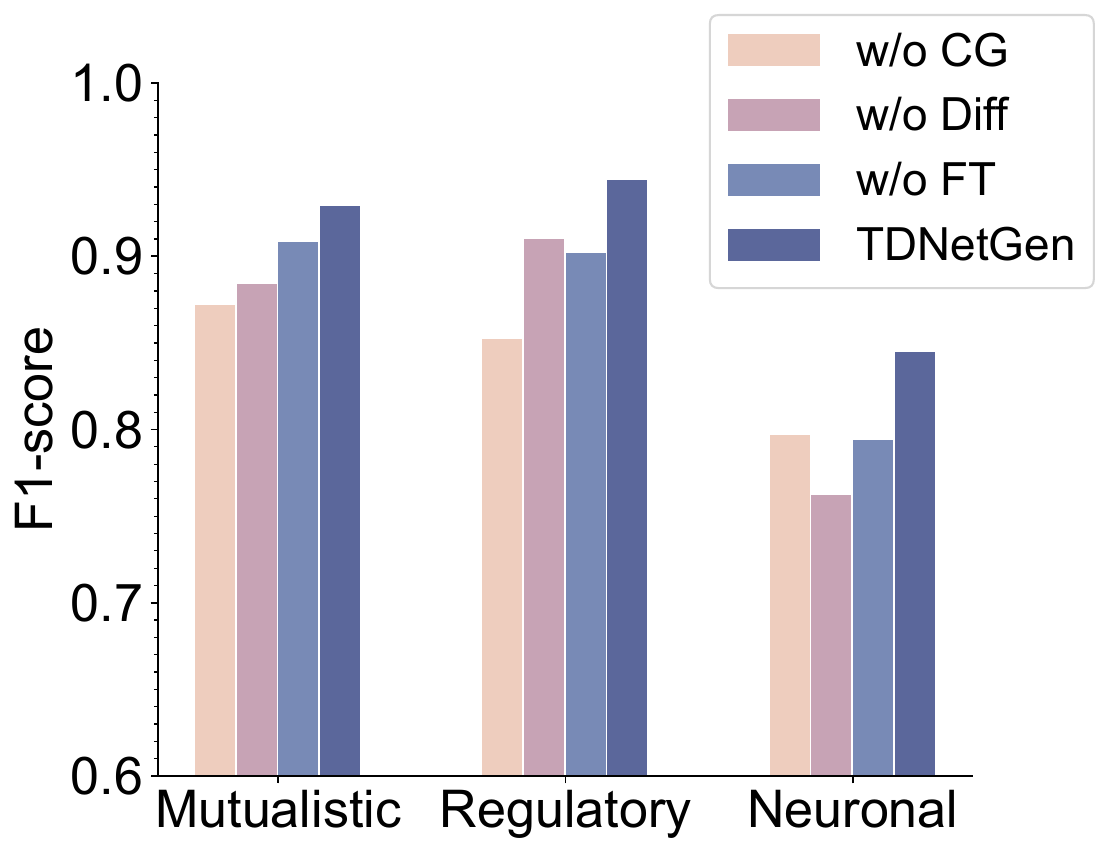}
    }
     \subfigure[ACC]{
    \includegraphics[width=0.475\linewidth]{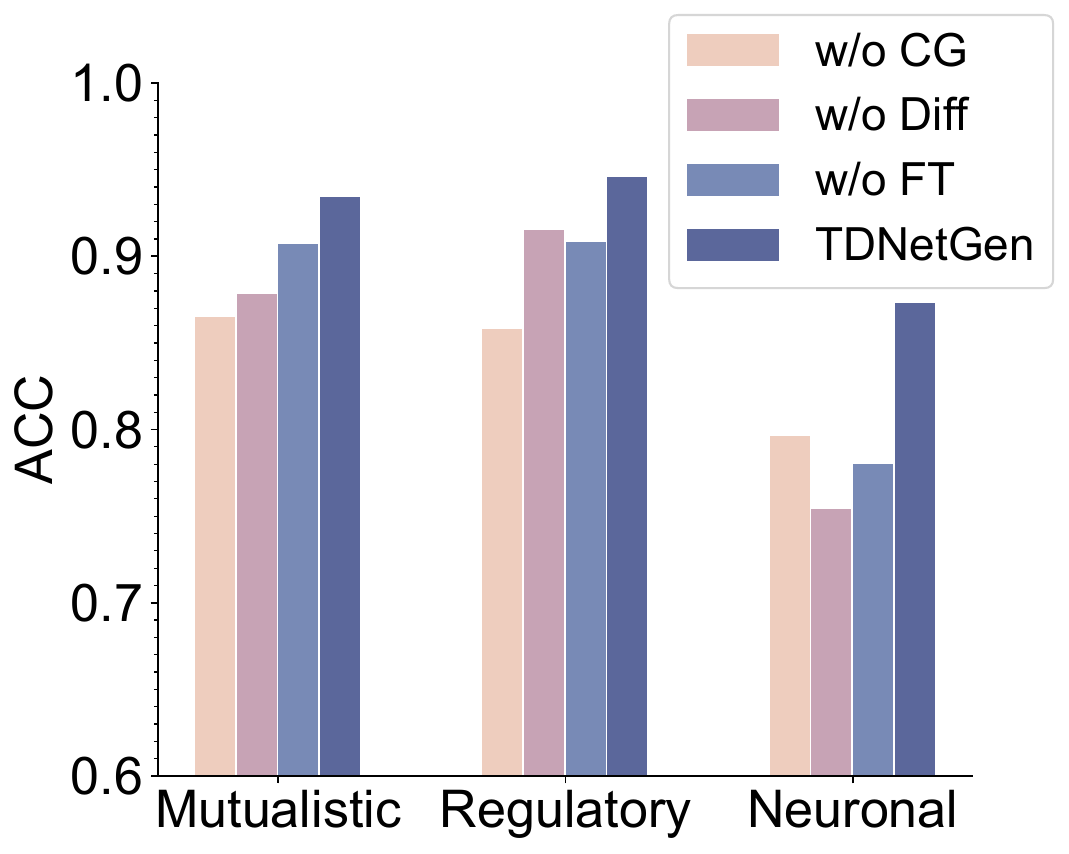}
    }
    \vspace{-0.6cm}
    \caption{Ablation studies on datasets. CG: Classifier guidance, Diff: Diffusion module, FT: Fine-tuning on dynamics learning module-produced trajectories.}
    \label{fig:ablation}
    \vspace{-0.7cm}
\end{figure}

To provide a comprehensive analysis and assess the effect of our designed components quantitatively, we conduct several ablation experiments via removing each design elements, and present the evaluation results in Figure~\ref{fig:ablation}.

\begin{itemize}[leftmargin=*,partopsep=0pt,topsep=0pt]
\item \textbf{Effectiveness of Classifier guidance.}
We first remove the design of classifier guidance and generate new network topologies via only unconditional topology diffusion module. The nodal state trajectories are simulated utilizing the dynamics learning module, and its resilience label is determined by the resilience predictor that has been trained on the labeled dataset. The results reveal that the F1-score of the ablation model significantly declines 
$6.14\%$, $9.75\%$, and $5.68\%$ 
compared to the full model design,
which underscores the importance of guided generation.
\item \textbf{Effectiveness of resilience predictor fine-tuning on dynamics learning module-produced trajectories.}
In this experiment, we remove the fine-tuning procedure of the resilience predictor on dynamics learning module-produced trajectories, instead utilizing the one trained with ground-truth trajectories. 
The results illustrate that fine-tuning could significantly enhance its guidance capabilities to generate higher-quality data, ultimately empowering the resilience predictor to be re-trained on it. 
\item \textbf{Architecture analysis.}
We compare our diffusion-based topology generation module with generative adversarial network (GAN) module. Specifically, we replace topology generation module as a GAN-based module proposed in~\cite{martinkus2022spectre}.
We use the topologies of unlabeled data to train the GAN model, and sample new topologies from it. The nodal state trajectories and the resilience label are produced by our dynamics learning module and the resilience predictor, respectively.
Experiments demonstrate that our original design of diffusion models exhibit superior generative performance compared to GANs,
which underscores the efficacy of diffusion models in capturing the underlying topology data distribution, thereby facilitating more accurate and reliable topology generation.

\end{itemize}
\vspace{-0.3cm}
\subsection{Augmentation with Limited Labels and Observations (RQ3)}
\begin{figure}[t!]
    \centering
    \subfigure[F1-score]{\includegraphics[width=0.475\linewidth]{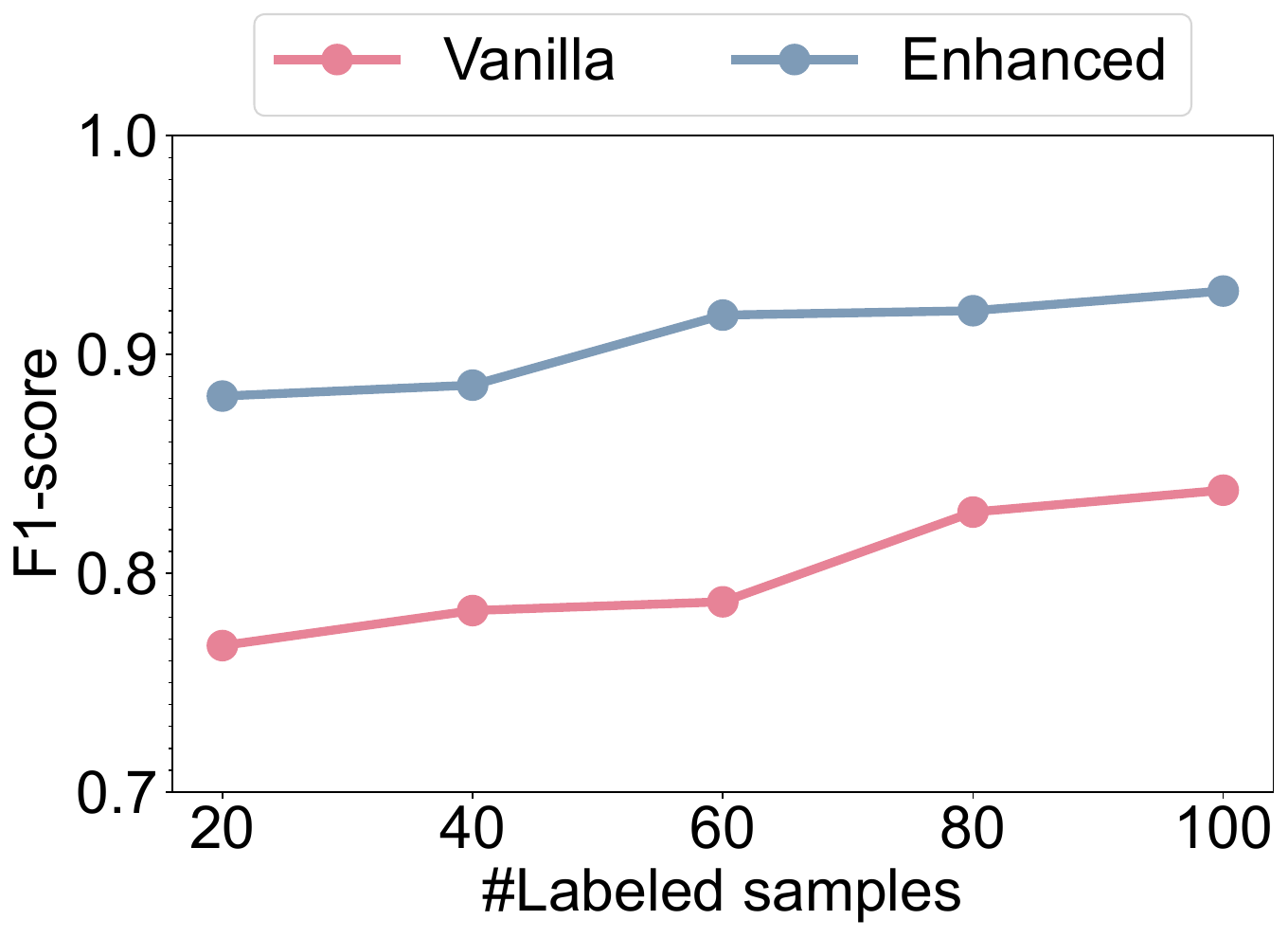}}
    \subfigure[ACC]{\includegraphics[width=0.475\linewidth]{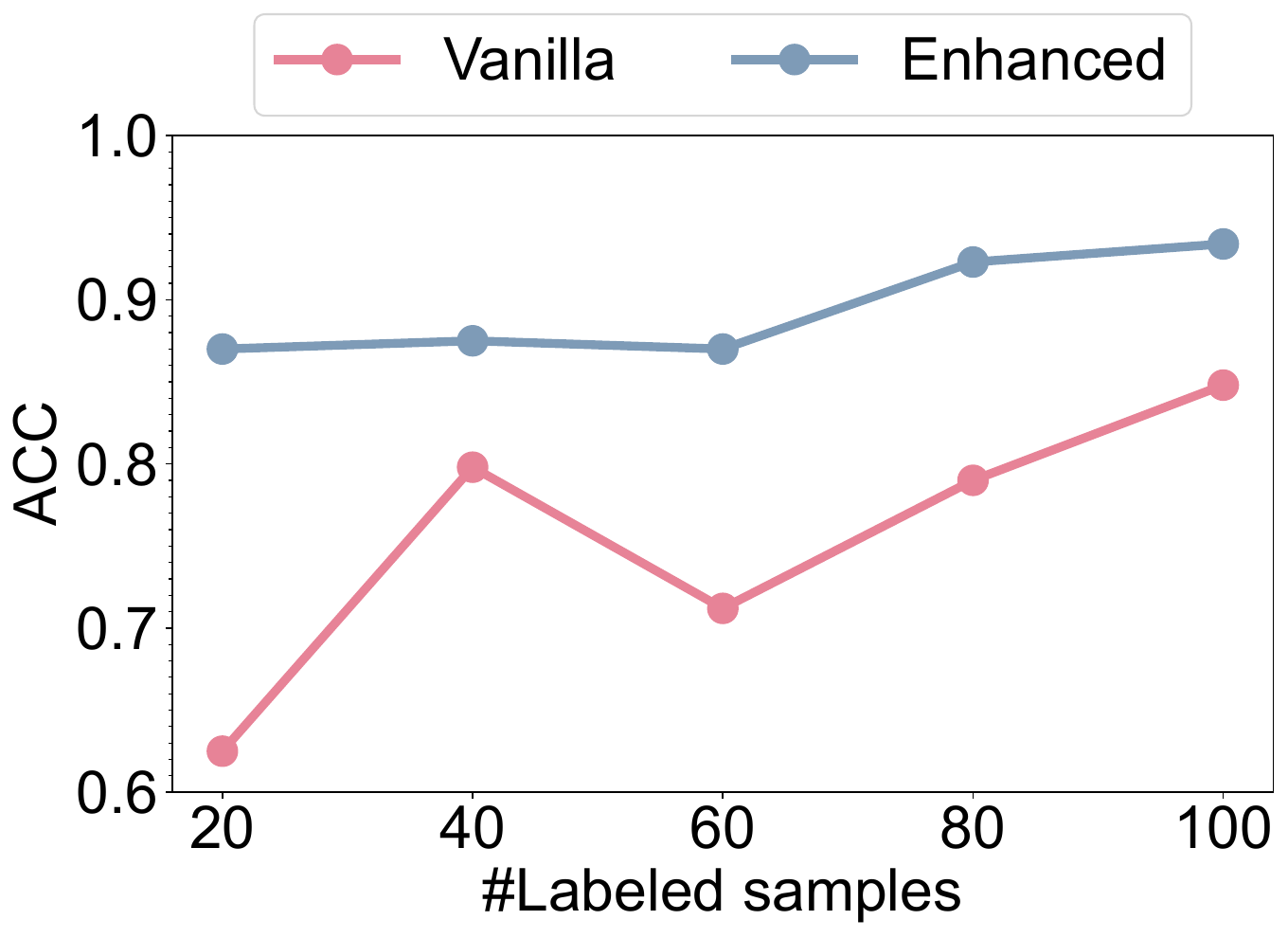}}
    \vspace{-0.4cm}
    \caption{Model performance with less labeled samples on mutualistic dataset.}
    \label{fig:mut-num}
    \vspace{-0.4cm}
\end{figure}

\begin{figure}[t!]
    \centering
    \subfigure[F1-score]{\includegraphics[width=0.475\linewidth]{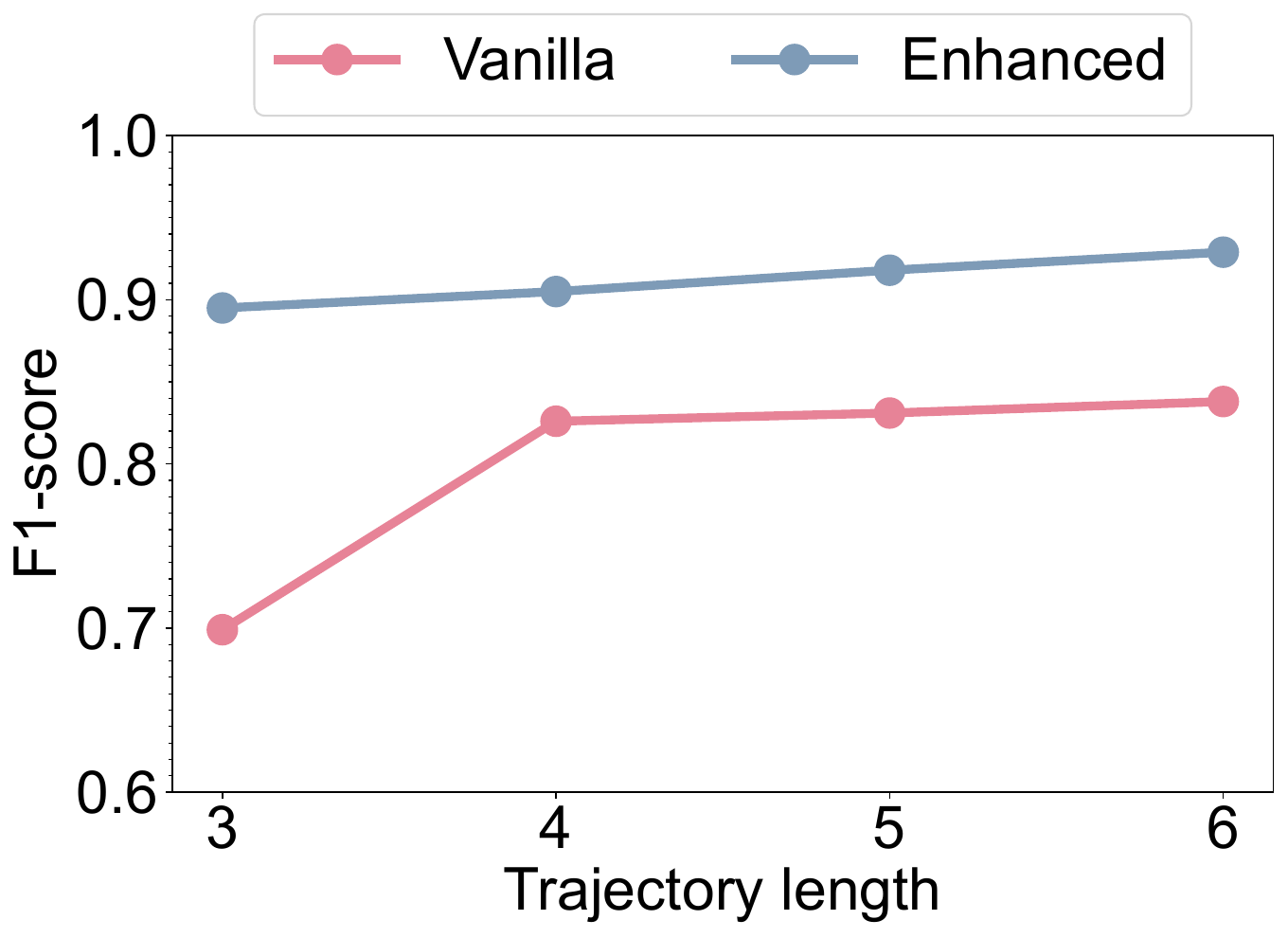}}
    \subfigure[ACC]{\includegraphics[width=0.475\linewidth]{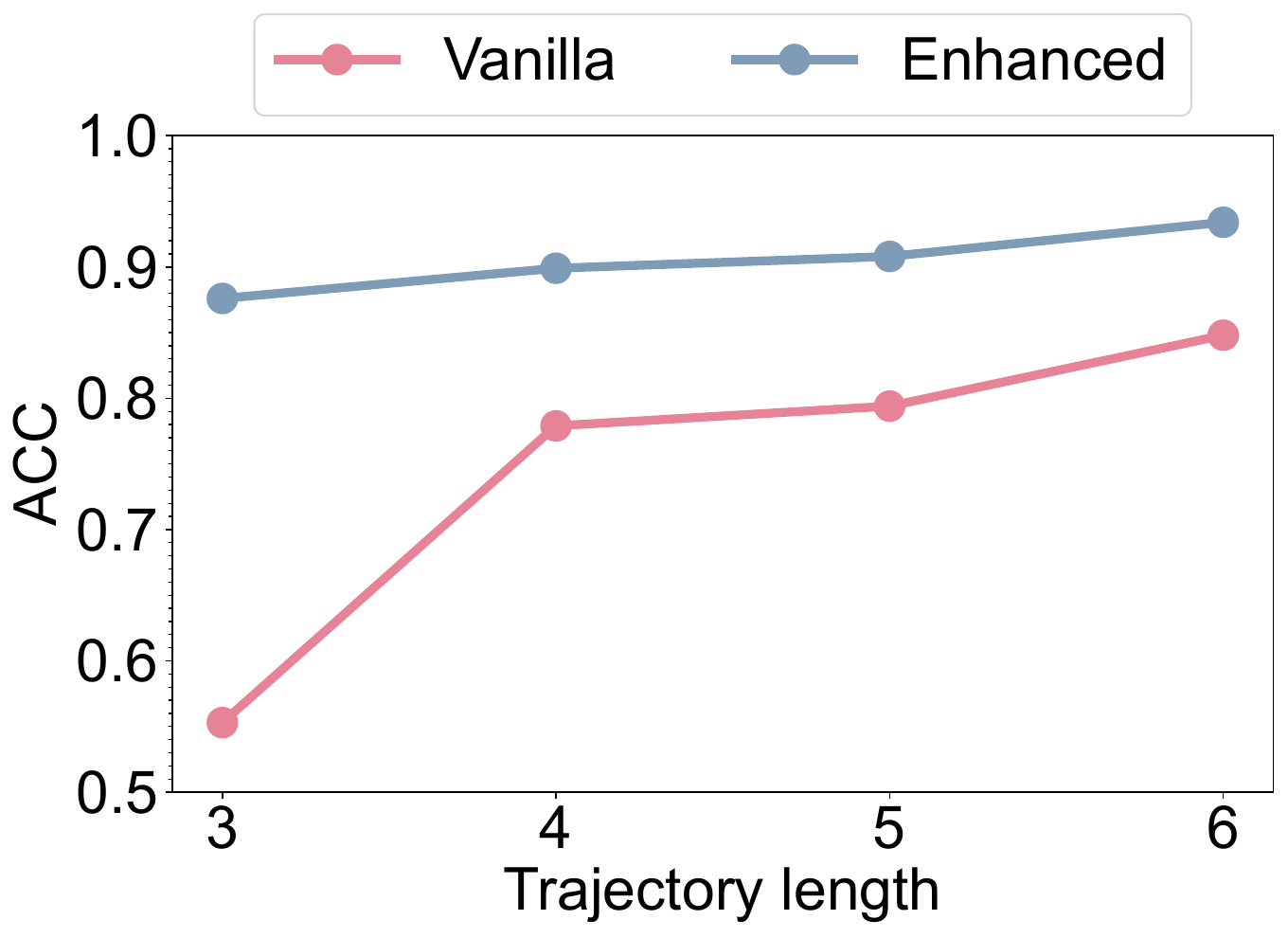}}
    \vspace{-0.4cm}
    \caption{Model performance with shorter nodal state trajectories on the mutualistic dataset.}
    \label{fig:mut-len}
    \vspace{-0.4cm}
\end{figure}

In this section, we investigate data augmentation capabilities of our proposed framework under conditions of more limited number of labeled samples and reduced observed trajectory lengths, representing more challenging scenarios. We illustrate the results on the mutualistic dataset in Figure~\ref{fig:mut-num}-\ref{fig:mut-len}.
\begin{itemize}[leftmargin=*,partopsep=0pt,topsep=0pt]
    \item \textbf{Less labeled samples.} 
    We investigate the performance of the vanilla model, where the numbers of labeled networks are in $\{20, 40, 60, 80, 100\}$, as well as the enhanced model trained on the augmented data generated by TDNetGen. From the results, we find that the predictive performance of the vanilla model is generally proportional to the number of labeled data used for training. TDNetGen is robust to the limitation of labeled data, which can still generate reasonable samples to benefit the predictive performance of the vanilla model. These findings underscore the versatility and potential of our proposed framework, particularly in scenarios characterized by a scarcity of labeled data, which constitutes a small portion of the available dataset. 
    \item \textbf{Shorter nodal state trajectories.}
    We also investigate the performance of the vanilla model and TDNetGen while using shorter nodal state trajectories, which contain $\{3, 4, 5, 6\}$ time steps. We discover that the performance of the vanilla model improves with the increase in trajectory length since the model can extract more knowledge about nodal state dynamics from data to make more accurate resilience predictions. In this scenario, TDNetGen can also help to augment the model's performance, which suggests that even in situations where nodal state trajectories are costly to acquire, our framework remains applicable and effective for data augmentation purposes of simultaneously generating plausible topologies and nodal state trajectories of complex networks.
\end{itemize}

\begin{table}[t]
\centering
\caption{\rev{Overall predictive performance of models on BA, $\mathbb{S}^1$, SBM, and brain networks. }}
\vspace{-0.3cm}
\label{tab:synthetic}
\resizebox{\linewidth}{!}{
\begin{tabular}{cccccccccc}
\hline
                                                                                  &               & \multicolumn{2}{c}{\textbf{BA}} & \multicolumn{2}{c}{\textbf{$\mathbb{S}^1$}} & \multicolumn{2}{c}{\textbf{SBM}} & \multicolumn{2}{c}{\textbf{Brain}} \\
\multicolumn{2}{c}{\# Training samples}                                                         & \multicolumn{2}{c}{100}                                                 & \multicolumn{2}{c}{100}                                                 & \multicolumn{2}{c}{100}  & \multicolumn{2}{c}{100}                                               \\ \hline
\multicolumn{2}{c}{\textbf{Model}}                                                              & \textbf{F1}                        & \textbf{ACC}                       & \textbf{F1}                        & \textbf{ACC}                       & \textbf{F1}                        & \textbf{ACC}      & \textbf{F1}                        & \textbf{ACC}                  \\ \hline
\multicolumn{2}{c}{Vanilla model}                                                                    & 0.814                              & 0.798                              & 0.776                              & 0.828                              & 0.767                              & 0.841 & 0.792 & 0.827                              \\ \hline
Self-training                                                                     & ST     & 0.767                              & 0.754                              & 0.774                              & 0.787                              & 0.778                              & 0.802    & 0.805                              & 0.796                          \\ \hline
\multirow{6}{*}{\begin{tabular}[c]{@{}c@{}}Self-supervised \\ learning\end{tabular}} & EdgePred~\cite{hu2019strategies}    & 0.797                              & 0.780                              & 0.747                              & 0.806                              & 0.784                              & 0.733         & 0.725 & 0.685                     \\
                                                                                  & AttrMask~\cite{hu2019strategies}    & 0.788                              & 0.776                              & 0.750                              & 0.805                              & 0.755                              & 0.760                              & 0.733 & 0.741\\
                                                                                  & ContextPred~\cite{hu2019strategies} & 0.792                              & 0.790                              & 0.771                              & 0.819                              & 0.754                              & 0.758                              &0.727 & 0.722\\
                                                                                  & InfoMax~\cite{velivckovic2018deep}& 0.776                              & 0.765                              & 0.784                              & {\ul 0.820}                        & 0.812                              & 0.833                              &0.743 & 0.775\\
                                                                                  & GraphLog~\cite{xu2021self}    & 0.783                              & 0.713                              & 0.780                              & 0.816                              & 0.759                              & 0.764                              & 0.745 & 0.757\\
                                                                                  & D-SLA~\cite{kim2022graph}       & 0.817                              & 0.823                              & 0.790                              & 0.813                              & 0.825                              & 0.820                              & 0.772 & 0.765\\ \hline
\multirow{2}{*}{\shortstack{Graph data \\ augmentation}}                                                              & TRY~\cite{gao2016universal}      & {\ul 0.837}                        & {\ul 0.840}                        & 0.791                              & 0.796                              & {\ul 0.855}                        & {\ul 0.858}   & 0.826 & 0.831                     \\
                                                                                  & G-Mixup~\cite{han2022g}     & 0.834                              & 0.837                              & {\ul 0.807}                        & 0.811                              & 0.852                              & 0.851                              & {\ul 0.839} & {\ul 0.844}\\ \hline
\multicolumn{2}{c}{\textbf{TDNetGen (w/o trajectories)}}                                        & \multicolumn{1}{l}{\textbf{0.842}} & \multicolumn{1}{l}{\textbf{0.846}} & \multicolumn{1}{l}{\textbf{0.823}} & \multicolumn{1}{l}{\textbf{0.830}} & \multicolumn{1}{l}{\textbf{0.886}} & \multicolumn{1}{l}{\textbf{0.890}} & \textbf{0.873} & \textbf{0.870} \\
\multicolumn{2}{c}{\textbf{TDNetGen}}                                                           & \textbf{0.870}                     & \textbf{0.850}                     & \textbf{0.856}                     & \textbf{0.875}                     & \textbf{0.935}                     & \textbf{0.937}  & \textbf{0.914} & \textbf{0.907}                    \\ \hline
\multicolumn{2}{c}{\textbf{Improvement}}                                                            & \textbf{3.99\%}                    & \textbf{1.19\%}                    & \textbf{6.07\%}                    & \textbf{6.71\%}                    & \textbf{9.36\%}                    & \textbf{9.21\%}  & \textbf{8.93\%} & \textbf{7.46\%}                  \\ \hline
\end{tabular}
}
\end{table}
\vspace{-0.3cm}
\subsection{Robustness against Different Network Types and Scales (RQ4)}
We consider other network models, including Barabási–Albert model~\cite{albert2002statistical}, $\mathbb{S}^1/\mathbb{H}^2$ model~\cite{serrano2008self}, and stochastic block model (SBM)~\cite{holland1983stochastic}, which have more complex and heterogeneous structural properties. Moreover, we also evaluate the scalability of our framework on large-scale empirical brain networks with 998 nodes in maximum. For each dataset, we obtain nodal states of networks via neuronal dynamics, and other experimental settings are the same as Section~\ref{subsec:expsetting}. The details of dataset construction are shown in Appendix~\ref{app:data}. We demonstrate the results in Table~\ref{tab:synthetic}, which indicates that our framework can still achieve the best augmentation performance on more broad types and scales of networks with complex structural properties and different network sizes.

\vspace{-0.1cm}
\section{Related Works}
\subsection{Resilience Prediction of Complex Networks}
Existing works on resilience prediction are mainly categorized to analytical estimations from physical theories~\cite{gao2016universal,laurence2019spectral, morone2019k}. Gao et al.~\cite{gao2016universal} propose to reduce the dimension of complex networks to single-parameter systems based on mean-field theory, thus we can easily analyze the equilibrium of 1-D ODE problem and predict the resilience of complex networks. Laurence et al.~\cite{laurence2019spectral} perform dimension reduction based on spectral graph theory on the dominant eigenvalues and eigenvectors of adjacency matrices.
Morone et al.~\cite{morone2019k} develop a resilience prediction methodology by quantifying the k-core structure within networks. Despite their effectiveness, they often pre-suppose a detailed understanding of nodal state dynamics, which is usually not available in practical scenarios. In our work, we design data-driven methods that extract topology and nodal state dynamics information from observational data, allowing for resilience predictions without the need for prior knowledge.

\subsection{Diffusion Models on Graphs}
Diffusion probabilistic models have been widely used in text, image, audio generation, etc.~\cite{austin2021structured,yang2023diffsound,luo2022antigen, yuan2024spatiotemporal}. Recently, some existing works have applied the diffusion model to the field of graph generation~\cite{vignac2022digress, huang2022graphgdp, tseng2023graphguide, chen2023efficient}. Huang et al.~\cite{huang2022graphgdp} define a stochastic differential equation (SDE) that smoothly converts graphs with complex distribution to random graphs, and samples new graphs by solving the reverse-time SDE. 
Tseng et al.~\cite{tseng2023graphguide} propose GraphGUIDE to achieve interpretable and controllable graph generation, wherein edges in graph are flipped or set at each discrete time step.
Chen et al.~\cite{chen2023efficient} propose to leverage graph sparsity during each step of diffusion process, which only focuses on a small portion of nodes and considers edge changes between them.
In contrast to existing contributions focused primarily on graph structures, our research extends to the generation of complex networks, which encompasses not merely the graph topology but also integrates nodal state trajectories, thereby facilitating the generation of comprehensive network data.
\subsection{Learning from Unlabeled Data}
Typical approaches of learning from unlabeled data for graph classification include pre-training on self-supervised tasks~\cite{xu2021self,kim2022graph}, self-training~\cite{iscen2019label, tagasovska2019single,amini2020deep,huang2022uncertainty}, and graph data augmentation~\cite{han2022g}.
Although pre-training proves to be effective for vision and language-related tasks, it can hardly help the resilience prediction task because of the disparity between hand-crafted and downstream prediction tasks~\cite{kim2022graph, inae2023motif}. Therefore, we still lack a universal self-supervised task that learns from unlabeled graphs and improves the performance of downstream scenarios.
Self-training tasks assign pseudo-labels to unlabeled data by leveraging the model itself, followed by the retraining of the model with pseudo-labeled data. Existing works~\cite{tagasovska2019single,amini2020deep,huang2022uncertainty} focus on uncertainty estimation of assigned labels to minimize the impact of noisy pseudo-labels. Furthermore, Liu et al.~\cite{liu2024data} learn data distributions from unlabeled graphs with diffusion models, and to generate task-specific labeled graphs for data augmentation. Compared with their work, our proposed TDNetGen framework considers more intricate scenarios of complex networks with interplay between topology and nodal state dynamics. Our framework can extract knowledge from full unlabeled complex network samples, thereby generating high-quality augmented data that benefits the training of prediction models.

\section{Conclusions}
In this work, we propose an effective framework, TDNetGen, for complex network resilience prediction. It not only addresses the problem in a data-driven manner without prior knowledge about groud-truth dynamics, but also solves labeled data sparsity problem with the generative augmentation of jointly modeling network topology and dynamics. Extensive experiments demonstrate the superiority of TDNetGen and also highlight its robustness within less labeled data and dynamics information conditions. The methodology introduced in this paper provides a novel perspective for improving resilience prediction through data augmentation, that is, leveraging the untapped potential of unlabeled data to enhance the learning process. 

\section*{Acknowledgment}
This work is supported in part by National Natural Science Foundation of China under U23B2030, 62272260, U21B2036.
\clearpage

\bibliographystyle{ACM-Reference-Format}
\balance
\bibliography{sample-base}

\appendix
\vspace{-0.3cm}
\section{Appendix}
\subsection{Details of Parameterization Network}\label{app:network}
\subsubsection{Graph Transformer}
The parameterization of the denoising network $h_\theta$ employs the Graph Transformer architecture~\cite{dhariwal2021diffusion}. 
The input consists of the noisy graph features, and the output is the edge distribution of clean graphs. Each layer of the Graph Transformer can be represented as follows:
\begin{gather}
h_i^{l+1}=O_h^l \|_{k=1}^K\left(\sum_{r_j \in \mathcal{N}_{r_i}} \alpha_{i j}^{k, l} V^{k, l} h_j^l\right), \\
e_{i j}^{l+1}=O_e^l \|_{k=1}^K\left(a_{i j}^{k, l}\right), \\
\alpha_{i j}^{k, l}=\operatorname{softmax}_j\left(a_{i j}^{k, l}\right), \\
a_{i j}^{k, l}=\left(\frac{Q^{k, l} h_i^l \cdot P^{k, l} h_j^l}{\sqrt{d_k}}\right)+W^{k, l} e_{i j}^l,
\end{gather}
where $h_i^l$ denotes the embedding of node $i$ of the $l$-th layer, $e_{ij}^l$ represents the embedding of edge connecting node $i$ and $j$. $O, Q, P, V, $ and $W$ with different superscripts are trainable parameters. $k= \{1, 2, \cdots, K\}$ denotes the attention heads, and $\|$ represents the concatenation operator. After the last layer of Graph Transformer, edge embeddings are fed into an MLP to predict the existence of edges $\{e_{ij}\}_{1 \leq i, j \leq N}$, where $N$ is the number of nodes in the network.
\subsubsection{Node features.}
We include node features in both structural and spectral domains to enhance the performance of the Graph Transformer, and choose the same features as in~\cite{vignac2022digress}. Specifically, structural features include the cycles, indicating \textit{i.e.,} how many $k$-cycles the node belongs to,  since message-passing cannot detect cycle structures~\cite{chen2020can}. For spectral features, we first compute the graph Laplacian then consider the number of connected components and the two first 
eigenvectors of the non-zero eigenvalues.
\vspace{-0.3cm}
\subsection{Details of Data Collection}\label{app:data}
\subsubsection{Mutualistic dynamics}
Nodal state trajectories of networks in mutualistic dataset are simulated via the following differential equations:
\begin{equation}
\frac{d x_i}{d t}=B+x_i\left(1-\frac{x_i}{K}\right)\left(\frac{x_i}{C}-1\right)+\sum_{j=1}^N A_{ij} \frac{x_i x_j}{D+E x_i+H x_j}.
\end{equation}
We use the fourth-order Runge-Kutta stepper, with a high initialization $\mathbf{x} = \mathbf{5}$ and a low initialization $\mathbf{x} = \mathbf{0}$, simulating two trajectories, which represent a thriving stable ecosystem and an ecosystem after a catastrophe, respectively. We set the terminal simulation time as $T_{max}=50$, and the interval $\Delta t=0.5$. 
Nodal states of networks with mutualistic dynamics can encounter a bifurcation~\cite{holland2002population}, transitioning from a resilient phase characterized by a single, desired high equilibrium $\mathbf{x}^H$ to a non-resilient phase with both the desired equilibrium $\mathbf{x}^H$ and the low equilibrium $\mathbf{x}^L$. 
We denote the averaged nodal states of the network from high and low initializations as $\langle \mathbf{x}^{(h)}\rangle$ and $\langle \mathbf{x}^{(l)}\rangle$; therefore, to define the resilience labels of networks, we compare $\langle \mathbf{x}^{(h)}\rangle$ and $\langle \mathbf{x}^{(l)}\rangle$ at the terminal time. If $|\langle \mathbf{x}^{(h)}\rangle-\langle \mathbf{x}^{(l)}\rangle| > r$, we conclude that the network cannot recover after perturbations and has two equilibrium $\mathbf{x}^H$ and $\mathbf{x}^L$, thus it is non-resilient. $r$ is a pre-defined threshold, and we set $r = 3.5$ in our experiments.
\subsubsection{Regulatory dynamics}
Nodal state trajectories of networks in regulatory dataset are simulated via the following differential equations:
\begin{equation} \label{equ:gene}
    \frac{d x_i}{d t}=-B x_i^f+\sum_{i=1}^N A_{i j} \frac{x_j^h}{x_j^h+1}.
\end{equation}
Similar to mutualistic dynamics, we use the fourth-order Runge-Kutta stepper, set the terminal simulation time $T_{max}=50$ and the interval $\Delta t = 0.5$. Regulatory dynamics in Equ.~(\ref{equ:gene}) has a trivial fixed point (as well as equilibrium) $\mathbf{x} = \langle \mathbf{x} \rangle = 0$, and for resilient networks with this dynamics, its nodal states have another equilibrium $\langle \mathbf{x} \rangle > 0$~\cite{alon2019introduction}. To avoid the fixed-point equilibrium, we randomly initialize the model with $\mathbf{x} = [1,5]$ and use the terminal nodal state $\langle \mathbf{x} \rangle$ to determine its resilience. Specifically, a network is deemed resilient if $\langle \mathbf{x} \rangle > 0$. Conversely, the non-resilient network can only converge to the equilibrium of $\langle \mathbf{x} \rangle = 0$.
\subsubsection{Neuronal dynamics}
Nodal state trajectories of networks in neuronal dataset are simulated with the following differential equations:
\begin{equation}
\frac{dx_i}{d t}=-x_i+\sum_{j=1}^N A_{i j} \frac{1}{1+e^{\mu-\delta x_j}}.
\end{equation}
The ODE solver, terminal, and interval time settings are the same as mutualistic and regulatory dynamics. 
Non-resilient networks exhibit either a bi-stable phase, wherein both a high equilibrium, denoted as $\mathbf{x}^{H}$, and a low equilibrium, denoted as $\mathbf{x}^{L}$, can exist, or only a single low equilibrium $\mathbf{x}^{L}$ exists. On the other hand, resilient neuronal networks are distinguished by their maintenance of a high equilibrium $\mathbf{x}^{H}$~\cite{wilson1972excitatory}. 
We initialize the nodal state with a high initialization $\mathbf{x} = \mathbf{5}$ and a low initialization $\mathbf{x} = \mathbf{0}$, simulating two trajectories. 
We compare $\langle \mathbf{x}^{(h)}\rangle$ and $\langle \mathbf{x}^{(l)}\rangle$ at the terminal time to define the resilience labels of networks. If $|\langle \mathbf{x}^{(h)}\rangle-\langle \mathbf{x}^{(l)}\rangle| > r$ or $\langle \mathbf{x}^{(h)}\rangle < m$ and $\langle \mathbf{x}^{(l)}\rangle < m$, we conclude that the network cannot recover after perturbations and have two equilibrium $\mathbf{x}^H$ and $\mathbf{x}^L$, thus it is non-resilient. Otherwise, the network is resilient. $r$ and $m$ are pre-defined thresholds, and we set $r = 3.5$ and $m = 3$ in our experiments.
\subsubsection{Barab\'asi-Albert (BA) model.} BA network starts with a small number of nodes, and at each time step, a new node with $m$ edges is added to the network. These $m$ edges link the new node to $m$ different nodes that have already present in network. The probability $\Pi$ that a new node will connect to an existing node $i$ is proportional to the degree $k_i$ of node $i$. Mathematically, $\Pi(k_i)=\frac{k_i}{\sum_j k_j}$, where the summation is over all existing nodes $j$ in the network. This means that nodes with higher degrees have a higher likelihood of receiving new links, leading to a “rich-get-richer” effect. The resulting network from the BA model exhibits a power-law degree distribution, $P(k)\sim k^{-\gamma}$, where $\gamma$ is typically in the range of 2 to 3. In our BA network datasets, each network consists of 100$\sim$200 nodes, and we set $m=4$.
\subsubsection{$\mathbb{S}^1$/$\mathbb{H}^2$ model.} It is also called hyperbolic geometric graph model. In this model, nodes are placed in a hyperbolic disk. Each node is assigned to a radial coordinate $r$ and an angular coordinate $\theta$. $r$ follows an exponential distribution capturing the heterogeneity of node degrees, while $\theta$ is uniformly distributed between $0$ and $2\pi$ representing the similarity or latent feature spaces of nodes. The connection probability of node $i$ and $j$ depends on their hyperbolic distance $d_{ij}=r_i+r_j+2\log (\sin(\frac{\theta_{ij}}{2}))$. Nodes are more likely connected if they are close in the hyperbolic space, reflecting the principle of preferential attachment and similarity. In our datasets, each network consists of 100$\sim$200 nodes, and the inverse temperature controlling the clustering coefficient $\beta$, the exponent of the power-law distribution for hidden degrees $\gamma$, and the mean degree of the network are set to $1.5$, $2.7$, and $5$, respectively.
\subsubsection{Stochastic block model (SBM)} In this model, nodes are partitioned into $K$ distinct groups. The probability of an edge existing between any two nodes depends solely on the groups to which these nodes belong. In our SBM datasets, each network consists of $[2,5]$ communities and $[20, 40]$ nodes (both sampled uniformly). The inter-community edge probability is $0.3$, and the intra-community edge probability is $0.05$.
\subsubsection{Empirical brain networks.} 
we employ an empirical brain networks~\cite{sanhedrai2022reviving, bullmore2009complex} with 998 brain regions (nodes), which represents the physical fiber bundle connections between them. The empirical network also has a natural modular structure owing to the brain's two hemispheresthis, indicating its complex structural properties. We generate 1900 topologies for unlabeled data, 100 topologies for labeled data, and 200 test data by randomly remove 0\%$\sim$15\% nodes from the empirical topology.
\vspace{-0.3cm}
\subsection{Details of Theoretical Baseline} \label{app:baseline}
In this section, we detail how we incorporate resilience theory from physics to provide insights on leveraging unlabeled data. 
\subsubsection{Gao-Barzel-Barab\'asi (GBB) theory}
From Gao-Barzel-Barab\'asi (GBB) theory~\cite{gao2016universal}, for a network with $N$ nodes, $N$-dimensional nodal state dynamics represented by Equ.~(\ref{equ:dynamics}) can be condensed to a 1-dimentional equation as:
\begin{equation}
    \frac{\mathrm{d} x_{\mathrm{eff}}}{\mathrm{d} t}=F\left(x_{\mathrm{eff}}\right)+\beta_{\mathrm{eff}} G\left(x_{\mathrm{eff}}, x_{\mathrm{eff}}\right), 
\end{equation}
\begin{align}
    x_{\mathrm{eff}} &=\frac{\mathbf{1}^T \mathbf{A}\mathbf{x}}{\mathbf{1}^T \mathbf{A}\mathbf{1}}=\frac{\langle \mathbf{s}^{out}\mathbf{x}\rangle}{\langle \mathbf{s}\rangle}, \\
    \label{equ:beta}
    \beta_{\mathrm{eff}} &= \frac{\mathbf{1}^T \mathbf{A} \mathbf{s}^{in}}{\mathbf{1}^T \mathbf{A}\mathbf{1}} = \frac{\langle \mathbf{s}^{out} \mathbf{s}^{in}\rangle}{\langle \mathbf{s}\rangle},
\end{align}
where $\mathbf{s}^{out} = (s_1^{out}, \cdots, s_N^{out})$ denotes the out-degrees of nodes and $\mathbf{s}^{in} = (s_1^{in}, \cdots, s_N^{in})$ denotes their in-degrees. $\langle \mathbf{s}^{out}\mathbf{x}\rangle = \frac{1}{N}\sum_{i=1}^N s_i^{out} x_i$ and $\langle \mathbf{s}\rangle =\langle \mathbf{s}^{in}\rangle = \langle \mathbf{s}^{out}\rangle$. Therefore, we can observe that network topology $\mathbf{A} \in \mathbb{R}^{N \times N}$ is condensed to a scalar $\beta_{\mathrm{eff}} \in \mathbb{R}$, which embeds the features of network topologies.
From dynamics equation, $f(\beta_{\mathrm{eff}}, x_{\mathrm{eff}}) = F\left(x_{\mathrm{eff}}\right)+\beta_{\mathrm{eff}} G\left(x_{\mathrm{eff}}, x_{\mathrm{eff}}\right)$, we can identify the bifurcation point of this dynamics, $\beta_{\text{eff}}^c$. If $\beta_{\text{eff}} < \beta_{\text{eff}}^c$, the undesired equilibrium will emerge, or the desired equilibrium will vanish. Otherwise, it has only one desirable equilibrium. Since \rev{$\beta_{\text{eff}}$} is calculated from network topology, we can conclude that a network with $\beta_{\text{eff}} < \beta_{\text{eff}}^c$ is non-resilient, and that of $\beta_{\text{eff}} > \beta_{\text{eff}}^c$ is resilient. Therefore, GBB theory provides an effective tool to predict network resilience. Its limitation is that precise analytical forms of $F(\cdot)$ and $G(\cdot)$ are required, which is hard to determine in real-world scenarios.
\subsubsection{Theory-guided data augmentation}\label{app:tryguided}
Here we discuss how to use GBB theory to perform data augmentation. For each network topology in labeled and unlabeled dataset, we calculate its $\beta_{\text{eff}}$ from Equ.~(\ref{equ:beta}). After that, we denote the \textit{minimum} $\beta_{\text{eff}}$ of resilient networks in the labeled dataset as $\beta^+$, and denote the \textit{maximum} $\beta_{\text{eff}}$ of non-resilient networks in the labeled dataset as $\beta^-$.
Therefore, for each network in the unlabeled dataset, if its $\beta_{\text{eff}} > \beta^+$, we label it as the resilient network; if its $\beta_{\text{eff}} < \beta^-$, we label it as the non-resilient network. Then the resilience predictor can further train on these newly-labeled data to enhance its predictive performance.
\vspace{-0.3cm}
\subsection{Additional Experiments}
\begin{figure}[t!]
    \centering
    \subfigure[F1-score]{\includegraphics[width=0.475\linewidth]{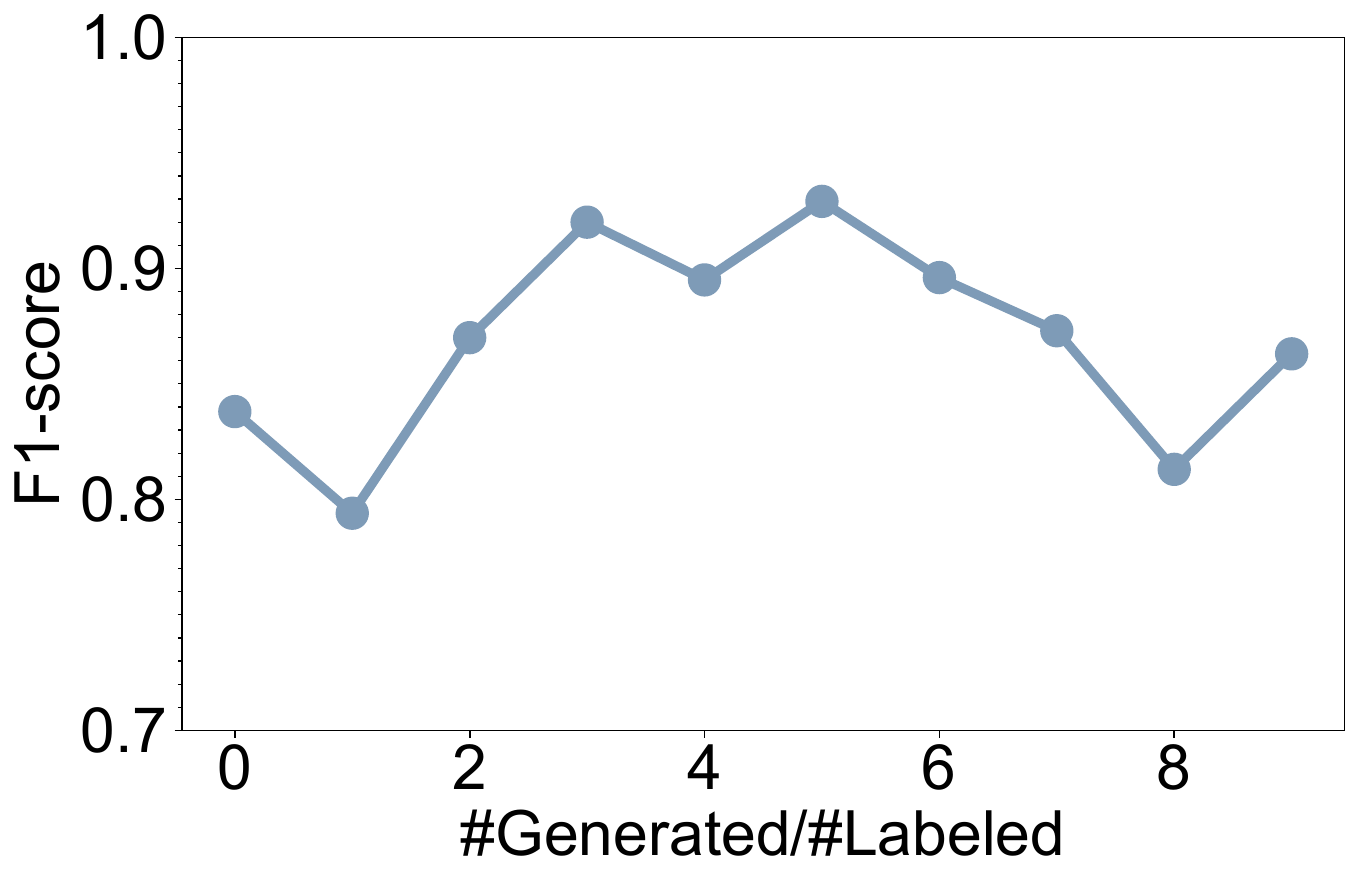}}
    \subfigure[ACC]{\includegraphics[width=0.475\linewidth]{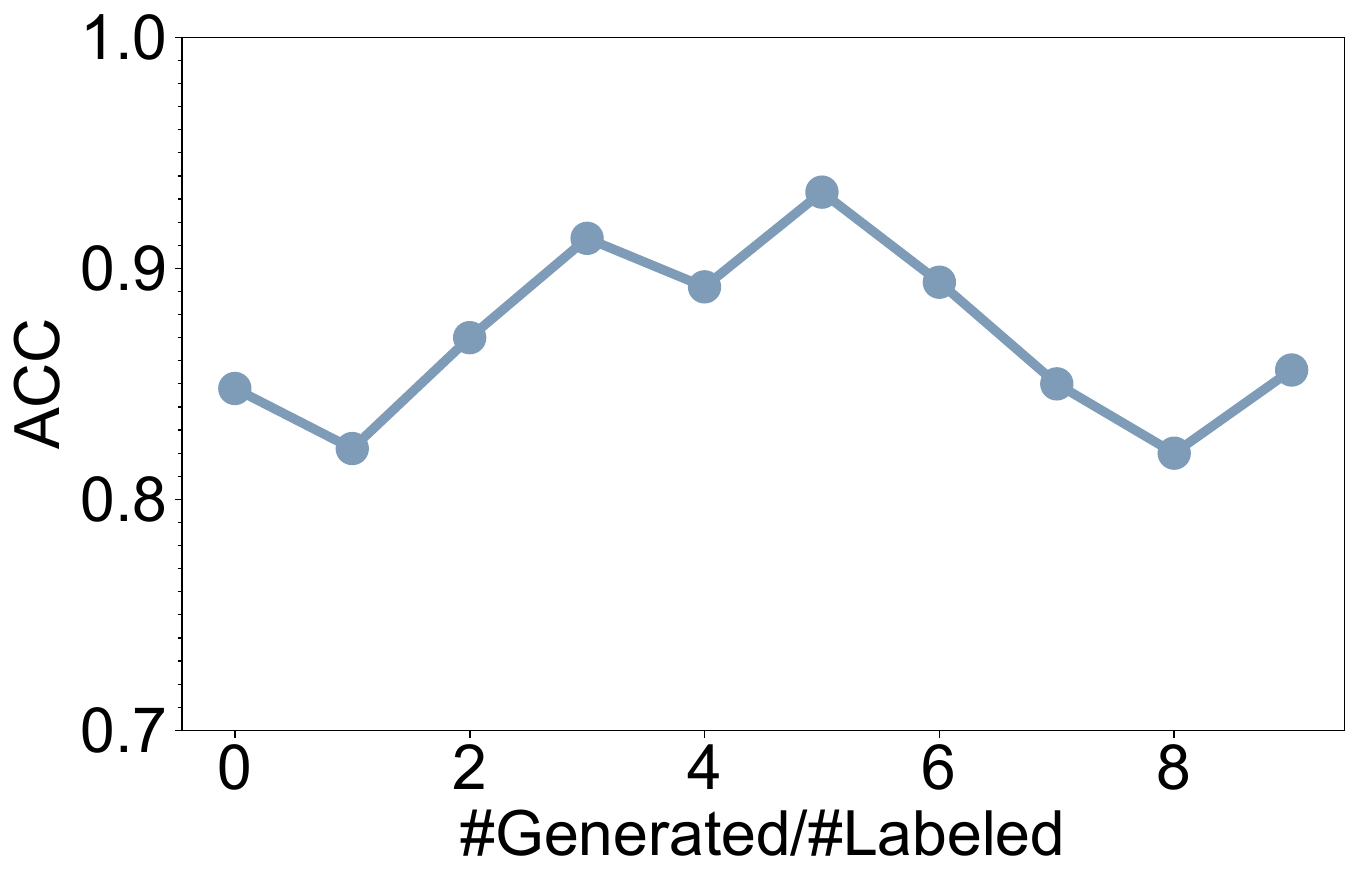}}
    \caption{Model performance with different number of generated samples on mutualistic dataset.}
    \label{fig:number}
\end{figure}

\noindent\textbf{Number of generated samples.}
We investigate the effect of used number of generated samples on the augmentation performance. The results on mutualistic dataset are shown in Figure~\ref{fig:number}, where the point in the 0 position of x-axis indicates the performance of the vanilla model. We find that there is an upper bound on the improvement introduced by data augmentation. Since we use the sub-optimal resilience predictor to guide the generation process, it is unavoidable to introduce generated data with fault labels. When the number of introduced generated data exceeds a threshold, the defect of noisy labels will exceed the positive effect of new training data, leading to the decrease of model performance.

\end{document}